\documentclass[11pt]{article}

\usepackage[preprint]{acl}
\usepackage{times}
\usepackage{latexsym}

\usepackage[T1]{fontenc}

\usepackage{microtype}

\usepackage{inconsolata}

\usepackage{graphicx}
\usepackage{enumitem}
\usepackage{amsmath}
\usepackage[ruled,vlined,linesnumbered,noend]{algorithm2e}
\SetInd{0.5em}{0.5em}
\DontPrintSemicolon
\SetKwInput{KwRequire}{Require}
\SetKwInput{KwEnsure}{Ensure}
\SetKwFor{ForAll}{for all}{do}{end for}
\usepackage{algpseudocode}
\usepackage{amssymb}
\usepackage{amsthm}
\usepackage[normalem]{ulem}
\usepackage{booktabs}
\usepackage{multirow}
\usepackage{graphicx}
\usepackage{needspace}

\usepackage{xcolor}
\usepackage{tcolorbox}
\tcbuselibrary{breakable}
\usepackage{graphicx}

\definecolor{venuebg}{HTML}{D4ECF8}
\definecolor{modelrow}{HTML}{F7F4E8}
\definecolor{oursrow}{HTML}{ECEBFF}

\newtcbox{\venuebox}{on line,
  boxrule=0pt,
  colback=venuebg,
  colframe=gray!15,
  arc=1.5pt,
  left=2pt,right=2pt,top=1pt,bottom=1pt,
  boxsep=0pt}

\usepackage[table]{xcolor}
\definecolor{oursrow}{HTML}{ECEBFF}
\definecolor{secondbest}{HTML}{E8F5E9}

\newtheorem{proposition}{Proposition}

\newcommand{\resultbest}[1]{%
  \colorbox{red!10}{$\mathbf{#1}$}}
\newcommand{\resultsecond}[1]{%
  \colorbox{orange!15}{\underline{$#1$}}}

\newcommand{\paskresultbox}[2]{\begingroup\setlength{\fboxsep}{0.4pt}\colorbox{#1}{\makebox[3.6em][c]{\rule[-0.4ex]{0pt}{1.9ex}#2}}\endgroup}
\renewcommand{\resultbest}[1]{\paskresultbox{red!10}{$\mathbf{#1}$}}
\renewcommand{\resultsecond}[1]{\paskresultbox{orange!15}{$\underline{#1}$}}

\usepackage{tikz}
\newcommand{\circnum}[1]{%
  \tikz[baseline=(char.base)]{
    \node[shape=circle, fill=black, text=white,
          inner sep=0.9pt, font=\scriptsize\sffamily\bfseries] (char) {#1};
  }%
}

\usepackage{etoolbox}
\AtBeginEnvironment{algorithmic}{\linespread{0.9}\selectfont}
\AtBeginEnvironment{algorithm}{%
  \setlength{\abovecaptionskip}{4pt}%
  \setlength{\belowcaptionskip}{2pt}}
\AtBeginDocument{%
  \setlength{\floatsep}{8pt plus 2pt minus 2pt}%
  \setlength{\textfloatsep}{12pt plus 2pt minus 4pt}%
  \setlength{\intextsep}{12pt plus 2pt minus 4pt}}

\title{Parser Transitions Already Reveal Structural Responsibility: KV Persistence for Structured Generation}

\author{
Linze Wu\quad Xinrui Chen\thanks{Corresponding author.}\\
Hangzhou Institute for Advanced Study,\\
University of Chinese Academy of Sciences, Hangzhou, China \\
\texttt{\{wulinze26, chenxinrui25\}@mails.ucas.ac.cn}
}

\begin{document}
\maketitle

\begin{abstract}
LLM agents use structured generation for JSON, SQL, and function calls. Constrained decoding ensures valid formats, but KV compression can discard history needed for correct function and argument selection. Structural responsibility can diverge from model saliency: importance scores can overlook generated tokens' structural roles. Parser transitions already computed for legality encode these roles, including required arguments and structural boundaries, providing a free retention signal. We introduce \textbf{PASK} (\underline{\textbf{P}}arser-\underline{\textbf{A}}ware \underline{\textbf{S}}tructural \underline{\textbf{K}}V Persistence), which reuses this signal for KV persistence by layer group without online importance estimation. Task-error sensitivity sets protection floors, while attention-output distortion ranks upgrades under the remaining budget. Offline calibration produces a frozen table for structural lookup and budget accounting. PASK outperforms budget-controlled baselines on function calling and text-to-SQL, improving BFCL accuracy by up to $8.67$ pp over the strongest comparator. Against Full KV, it cuts peak GPU memory by $47.5\%$ and achieves up to $3.3\times$ TPOT speedup. 
\end{abstract}

\section{Introduction}
\label{sec:introduction}

LLM agents access tools and databases through structured outputs: JSON, SQL~\citep{NEURIPS2023_83fc8fab}, function calls~\citep{pmlr-v267-patil25a}, and other machine-executable formats~\citep{geng2025jsonschemabenchrigorousbenchmarkstructured}. Constrained decoding ensures format validity by allowing only tokens consistent with a schema, grammar, or function signature~\citep{scholak-etal-2021-picard,geng-etal-2023-grammar,MLSYS2025_5c20ca4b}. Compressing the key--value (KV) cache of reusable token representations improves serving efficiency but can discard essential history: \textit{\textbf{grammatically valid calls may still select incorrect functions or arguments}}.

\begin{figure}[t]
\centering
\includegraphics[width=\columnwidth]{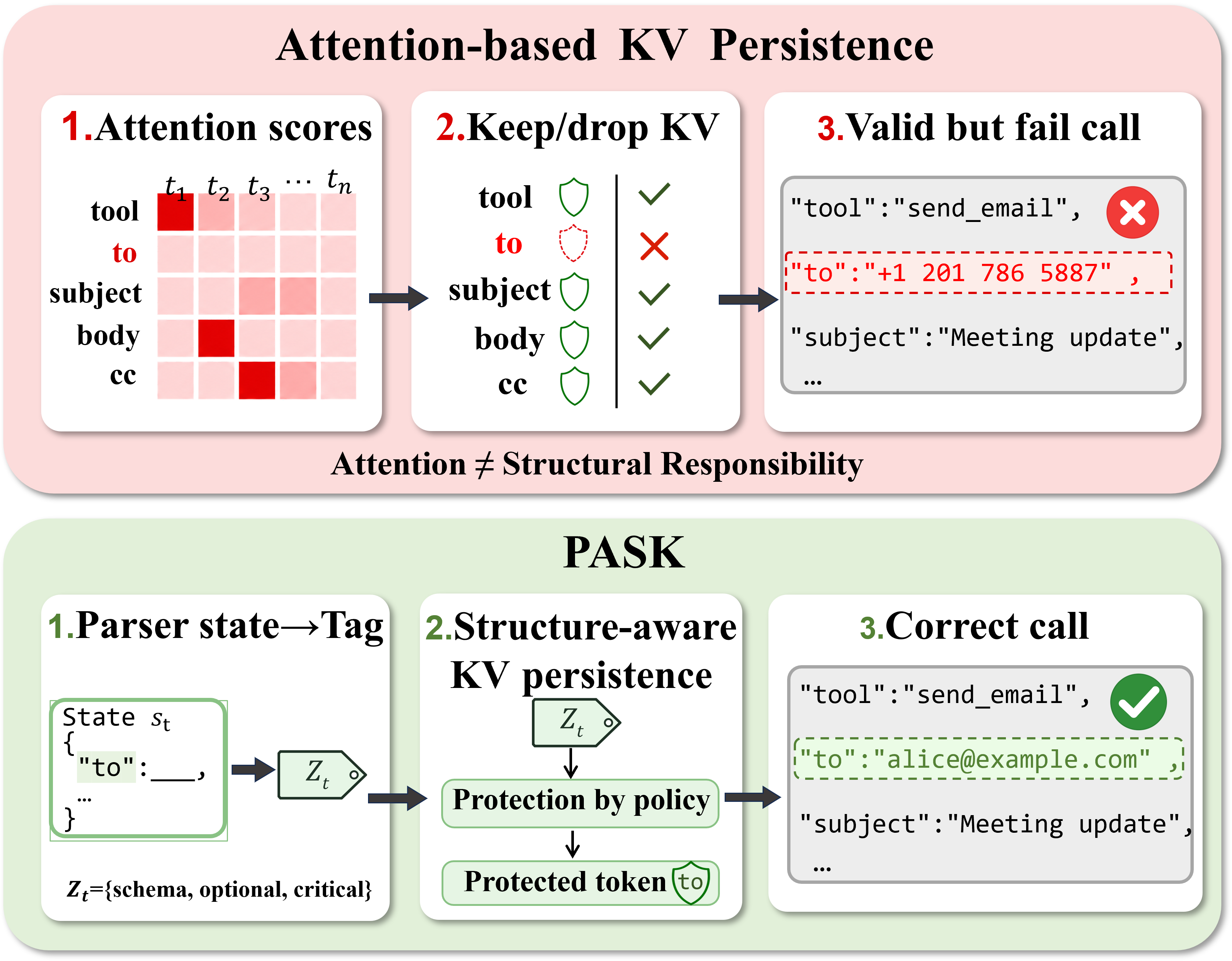}
\vspace{-0.3in}
\caption{Motivation: parser transitions expose structural roles complementing model-side saliency.}
\label{fig:pask-motivation}
\vspace{-0.1in}
\end{figure}

Existing KV policies guide retention using attention saliency~\citep{NEURIPS2023_6ceefa7b,NEURIPS2024_28ab4182}, quantization error~\citep{pmlr-v235-liu24bz,NEURIPS2024_028fcbcf}, recency~\citep{ICLR2024_5e5fd18f}, or generic token importance~\citep{zhou-etal-2025-dynamickv,wu-etal-2025-scope}. Structure-aware approaches use coarse roles in schema-dense inputs~\citep{mandal2026adaptivefilteringkvcache}. These signals can overlook generated tokens' structural roles, creating a responsibility--saliency mismatch between model importance and retention needs for task correctness. Under our shared constrained-decoding protocol at approximately one-third KV budget, attention-guided baselines lose up to $70.0$ accuracy points relative to Full KV on BFCL non-Live parallel-multiple calls. Fig.~\ref{fig:pask-motivation} motivates retention guided by structural responsibility.

Our insight is that constrained decoding already provides structural information for KV retention. As each token is accepted, the parser updates its state to track progress through the output structure. These transitions reveal roles such as required arguments, optional fields, and structural boundaries. \textit{\textbf{Already computed for legality enforcement, they provide a freely available signal for memory management.}} We repurpose this signal to connect a token's structural role to its KV persistence, giving the constraint engine a second role in protecting history for subsequent generation.

We introduce \textbf{PASK} (\underline{\textbf{P}}arser-\underline{\textbf{A}}ware \underline{\textbf{S}}tructural \underline{\textbf{K}}V Persistence), which assigns release, low-precision retention, or high-precision retention by parser-derived structural tag and Transformer layer group, allowing each token's protection to vary with depth. \textit{\textbf{Its risk-constrained allocation separates task protection from residual spending}}: offline interventions measure task-error sensitivity to releasing or quantizing each structure--layer group to set protection floors, while offline attention-output distortion ranks upgrades by distortion reduction per unit of additional memory. After development-set selection of the lowest-cost candidate satisfying a task-error tolerance, PASK compiles actions, floors, and upgrade priorities into a frozen table. During decoding, PASK retrieves them by structural tag and layer group and adjusts retention to the budget without recomputing importance or distortion. With fixed layer groups, per-token control work is independent of retained-history length. A lightweight prefill companion compresses prompt KV, enabling end-to-end compression of prompt and generated history.

PASK achieves SOTA accuracy among the evaluated budget-controlled KV compression methods on BFCL and BIRD. At roughly one-third total KV budget, it stays within $2.5$ points of Full KV on BFCL non-Live parallel-multiple calls across three models. Relative to Full KV, it \textit{\textbf{reduces peak GPU memory by $47.5\%$ and achieves up to $3.3\times$ TPOT speedup}} in the primary serving matrix. Ablations and transfer experiments support the design and calibration reuse.

Our contributions are:
\begin{itemize}[leftmargin=*, nosep]
\item We identify a responsibility--saliency mismatch in compressed structured generation and establish existing parser transitions as a structural signal for generated-KV retention.
\item We formulate risk-constrained KV persistence: task-error sensitivity sets protection floors; numerical distortion allocates remaining memory.
\item We develop PASK, compiling structural and layer-dependent retention decisions into a frozen policy with lightweight online control.
\item We demonstrate accuracy and serving gains on structured-generation benchmarks, with ablations and transfer experiments supporting the mechanism and policy reuse.
\end{itemize}

\section{Related Work}
\label{sec:related-work}
\textbf{Structured Generation.}
Structured generation enforces schemas, grammars, or function signatures so that outputs remain parseable and executable~\citep{scholak-etal-2021-picard,geng-etal-2023-grammar}. Systems compile them into guided decoding procedures~\citep{willard2023efficientguidedgenerationlarge,NEURIPS2024_724be447,beurerkellner2024guidingllmsrightway,ICLR2025_3d3a9e08}, explicit parser states~\citep{scholak-etal-2021-picard,MLSYS2025_5c20ca4b,li2026xgrammar2dynamicefficientstructured}, or token masks~\citep{MLSYS2025_5c20ca4b,li2026xgrammar2dynamicefficientstructured}; evaluations emphasize coverage, correctness~\citep{NEURIPS2023_83fc8fab,geng2025jsonschemabenchrigorousbenchmarkstructured,pmlr-v267-patil25a,chen-etal-2025-acebench}, and efficiency. PASK extends parser state usage from legality enforcement to KV retention.

\textbf{KV Compression.}
KV compression reduces inference memory through low-bit quantization~\citep{pmlr-v235-liu24bz,NEURIPS2024_028fcbcf,MLSYS2025_fbe2b2f7,sharma-etal-2025-minikv,boroujeni2026dontwastebitsadaptive} to store KV states with reduced numerical precision, attention- or distortion-based token selection~\citep{NEURIPS2023_a452a7c6,NEURIPS2023_6ceefa7b,ICLR2024_5e5fd18f,NEURIPS2024_28ab4182,ICLR2024_639a9a17,pmlr-v267-behnam25a,goel2025caotekvcacheselection,pmlr-v306-feng26j,zhang2026rdkvratedistortionbitallocation,jo-etal-2026-fastkv,chen-etal-2026-structkv} to preserve history estimated to matter for subsequent decoding, and system-level cache organization~\citep{kwon2023efficientmemorymanagementlarge} to reduce fragmentation and improve memory access efficiency. Complementary work exploits inter-layer redundancy~\citep{NEURIPS2024_fd070571}, non-uniform layer budgets~\citep{cai2025pyramidkvdynamickvcache}, task- and context-adaptive allocation~\citep{zhou-etal-2025-dynamickv,liu2026agentkv,pmlr-v306-chen26cd}, distillation~\citep{jiang2026kvsculptkvcachecompression}, reasoning-stage signals~\citep{NEURIPS2025_57e0358a,ICLR2026_b2b502c3}, and agent-level cache directives~\citep{ma2026leylinekvcachedirectives}. KV utility varies across tokens, layers, and tasks, but retention signals remain model-internal statistics or task-level heuristics that do not directly encode whether accepted generated tokens realize required keys, arguments, enum values, or schema boundaries.

\textbf{Structure-Aware Memory.}
PASK's closest antecedents incorporate temporal~\citep{shen2026triaxialkvextremelowprecisionkvcache}, semantic~\citep{basu2026transactionalattentionsemanticsponsorship,NEURIPS2025_2987f911,li2026intentkvcrossturnintentawarekv}, program-context~\citep{chen2026codecompstructuralkvcache}, and structural-role signals~\citep{mandal2026adaptivefilteringkvcache} into KV allocation or retention. Prior work exploits structural cues from attention patterns, program context, execution traces, or prompt organization. PASK instead repurposes token-level parser transitions already computed for legality during constrained decoding. This  signal exposes generated tokens' structural responsibility and guides their KV retention without online importance estimation.

\begin{figure*}[t]
    \centering
    \includegraphics[width=\textwidth]{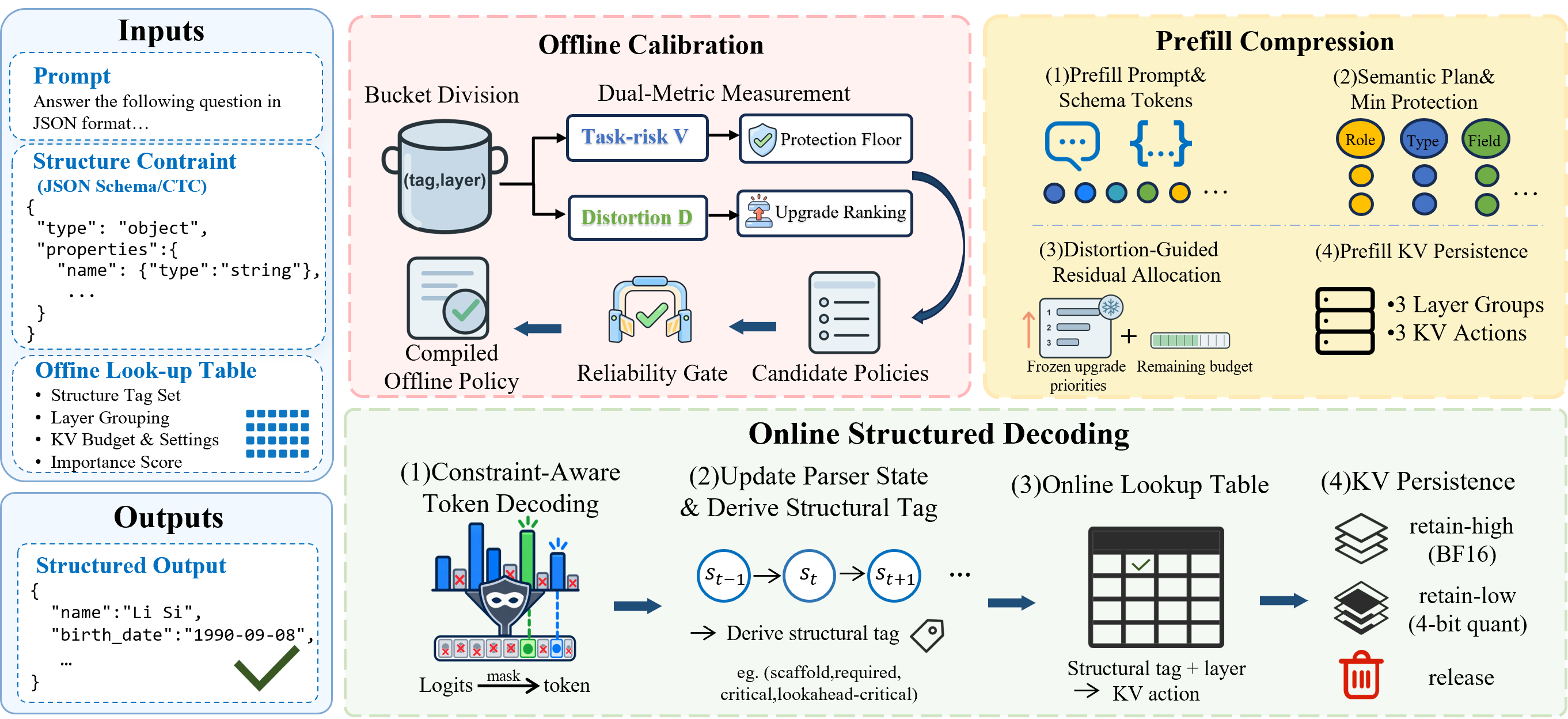}
    \vspace{-0.3in}
    \caption{Overview of PASK. Parser transitions and layer groups index an offline-calibrated table that assigns high-precision retention, low-precision retention, or release to generated KV states.}
    \label{fig:pask-overview}
    \vspace{-0.1in}
\end{figure*}

\section{Method}
\label{sec:method}
PASK reuses parser transitions computed for constrained decoding for layer-group KV persistence. Offline calibration separates task protection from distortion-based allocation, compiling a frozen table for online lookup and budget accounting.

\subsection{Preliminaries}
\label{sec:preliminaries}
\paragraph{KV cache.}
For one Transformer layer and head, causal attention at sequence position $i$ computes
\begin{equation}
\mathbf o_i=\mathbf V_i\,\mathrm{softmax}\!\left(\mathbf K_i^\top\mathbf q_i/\sqrt{d_k}\right),
\label{eq:prelim-attention}
\end{equation}
where $\mathbf q_i$ is the query, $\mathbf K_i=[\mathbf k_1,\ldots,\mathbf k_i]$ and $\mathbf V_i=[\mathbf v_1,\ldots,\mathbf v_i]$ contain keys and values, and $d_k$ is the key dimension. KV caching reuses past keys and values. Eviction renormalizes attention over retained entries; quantization changes their representations. Both can alter attention outputs.

\paragraph{Constrained decoding.}
Given request $r=(x,G)$ with prompt $x$ and constraint $G$, parser state $s_{t-1}$ summarizes $y_{1:t-1}$. Decoding selects $y_t\in\mathcal A_{\mathrm{legal}}(s_{t-1})$ and updates $s_t=\delta_G(s_{t-1},y_t)$ before the next forward pass computes the non-stop token's KV. Format validity does not ensure correct functions, arguments, or field values.

\subsection{Risk-Constrained KV Persistence}
\label{sec:method-setup}
For generated KV (prompt compression: Section~\ref{sec:prefill-companion}), PASK partitions Transformer layers into groups $\mathcal C$ and assigns each non-stop token $a_{t,c}\in\{R,L,H\}$ per group $c$: release, INT4 retention, or BF16 retention, ordered by increasing protection and cost. Recent tokens remain high-precision; releases occur after window exit. Scheduled long-range payload $m(a,c)$ includes group size and quantization scales, with $m(R,c)=0$. Recent-window overhead and physical GPU memory are accounted for separately (Appendix~\ref{app:pask-accounting}). Define $C_{\mathrm{full}}=\sum_c m(H,c)$ and
\begin{equation}
\begin{aligned}
R_{\mathrm{task}}(\pi)&=\mathbb E_{r\sim\mathcal D}\big[\mathbb I[\mathrm{Fail}(y_r^\pi)]\big],\\
C_{\mathrm{KV}}(\pi)&=\frac{\mathbb E_{r\sim\mathcal D}\big[\sum_{t=1}^{T_r^\pi}\sum_c m(a^\pi_{r,t,c},c)\big]}{\mathbb E_{r\sim\mathcal D}[T_r^\pi]C_{\mathrm{full}}},
\end{aligned}
\label{eq:risk-cost}
\end{equation}
where $T_r^\pi$ counts generated non-stop tokens and $\mathrm{Fail}$ follows the benchmark task oracle, including incomplete outputs. The denominator uses the same generated lengths. Relative to the all-high reference $\pi_{\mathrm{ref}}$, our objective is
\begin{equation}
\begin{aligned}
\pi^\star_\alpha&\in\arg\min\nolimits_{\pi\in\Pi_{\mathrm{struct}}}C_{\mathrm{KV}}(\pi)\\
&\mathrm{s.t.}\quad R_{\mathrm{task}}(\pi)-R_{\mathrm{task}}(\pi_{\mathrm{ref}})\leq\alpha.
\end{aligned}
\label{eq:main-objective}
\end{equation}
$\Pi_{\mathrm{struct}}$ contains policies using structural tags, layer groups, and request-local budget state, and $\alpha$ bounds the absolute increase in task failure probability. PASK searches a finite family, enforcing the tolerance empirically on development data.

\subsection{Parser-Derived Persistence Buckets}
\label{sec:persistence-buckets}
A deterministic grammar adapter maps each accepted transition, already computed for legality, to a structural tag and pairs it with a layer group:
\begin{equation}
\begin{aligned}
z_t&=\phi_G(s_{t-1},y_t,s_t)\\
&=(\mathrm{class}_t,\mathrm{role}_t,\mathrm{state}_t,\mathrm{next}_t),\\
q_{t,c}&=(z_t,c),\qquad c\in\mathcal C.
\end{aligned}
\label{eq:structural-bucket}
\end{equation}
The components identify token class, schema responsibility, grammar status, and an approaching critical region. Required arguments and field boundaries receive corresponding roles; $\mathrm{next}_t$ uses the current grammar frontier, not future tokens. These labels identify structural responsibilities, while offline task feedback determines their protection requirements. Buckets share calibrated decisions across token identities without online importance estimation. Unresolved tags default to high protection. Appendix~\ref{app:policy-construction} specifies the vocabulary, and Appendix~\ref{app:backend} describes parser integration. Lower, middle, and upper layer groups allow identical roles to receive different protection across depth~\citep{NEURIPS2024_fd070571,cai2025pyramidkvdynamickvcache,zhou-etal-2025-dynamickv}. For example, the same argument token may receive low-precision retention in early layers and high-precision retention in later layers.

\subsection{Offline Calibration and Compilation}
\label{sec:offline-calibration}
Task-error sensitivity determines minimum protection; attention-output distortion guides upgrades. Both offline measurements share decoding settings, prompt policy, and recent-window rules.

\paragraph{Task-error sensitivity.}
Policy $\pi_{q,a}$ applies $a\in\{R,L\}$ to every occurrence of bucket $q$ during complete autoregressive generation, keeping other buckets at $H$. Occurrences follow the intervened trajectory, allowing the measurement to capture downstream changes in generation. On the same calibration requests,
\begin{align}
&V(q,a)=\widehat R_{\mathrm{cal}}(\pi_{q,a})-\widehat R_{\mathrm{cal}}(\pi_{\mathrm{ref}}),\notag\\
&\bar V(q,H)=0,\qquad\bar V(q,L)=\max\{0,V(q,L)\},\notag\\
&\bar V(q,R)=\max\{\bar V(q,L),V(q,R)\}.
\label{eq:task-sensitivity}
\end{align}
The envelope conservatively orders empirical sensitivities by protection. For $0\leq\tau_{\mathrm{low}}\leq\tau_{\mathrm{high}}$,
\begin{equation}
a_{\min}^{\tau}(q)=
\begin{cases}
H,&\bar V(q,L)>\tau_{\mathrm{high}},\\
R,&\bar V(q,R)\leq\tau_{\mathrm{low}},\\
L,&\text{otherwise}.
\end{cases}
\label{eq:protection-floor}
\end{equation}
A bucket with modest attention distortion can therefore remain protected if releasing it exceeds the task-error threshold. Bucket thresholds construct local floors; $\alpha$ governs complete-policy selection, which checks joint effects.

\paragraph{Attention-output distortion.}
Using fixed reference queries and K/V tensors, we locally recompute attention after releasing or quantizing eligible entries in $q$. Reference tokens and queries remain unchanged, isolating numerical deviation from the task-level effects measured by $V$. Let $\mathcal S(q)$ contain affected reference query positions $(r,u)$, shared across actions. Define accumulated distortion per calibration request as
\begin{equation}
\begin{aligned}
D(q,a)&=\frac{1}{|\mathcal D_{\mathrm{cal}}|}\sum_{(r,u)\in\mathcal S(q)}\sum_{\ell\in c}\\
&\qquad\max_h\left\|o^{\mathrm{ref}}_{\ell,r,u,h}-o^{a(q)}_{\ell,r,u,h}\right\|_2^2,
\end{aligned}
\label{eq:distortion}
\end{equation}
where $h$ indexes query heads and $D(q,H)=0$. For adjacent upgrades $R\to L$ or $L\to H$,
\begin{equation}
\rho_D(q,a\to a')=\frac{[D(q,a)-D(q,a')]_+}{\bar N_q[m(a',c)-m(a,c)]},
\label{eq:upgrade-score}
\end{equation}
where $\bar N_q$ is the mean bucket count per reference request. Because a bucket upgrade affects every occurrence, the denominator measures its mean total incremental payload per request. Calibration details appear in Appendix~\ref{app:pask-calibration-details}.

\paragraph{Policy selection.}
For each threshold pair $\tau\in\mathcal T$, PASK starts from its floors and repeatedly applies the eligible upgrade with largest $\rho_D$, generating a nested candidate path. Each candidate retains its floor map and takes its nominal decode target $B_{\mathrm D}$ from its reference-trajectory payload ratio. We construct its reserve and evaluate the complete controller (Section~\ref{sec:online-lookup}) autoregressively on development requests, capturing joint task effects and budget-driven action changes:
\begin{equation}
\begin{aligned}
i^\star&\in\arg\min_{i\in\mathcal I_\alpha}\widehat C_{\mathrm{KV,dev}}(\pi_i),\\
\mathcal I_\alpha&=\{i:\widehat R_{\mathrm{dev}}(\pi_i)-\widehat R_{\mathrm{dev}}(\pi_{\mathrm{ref}})\leq\alpha\}.
\end{aligned}
\label{eq:dev-selection}
\end{equation}
Costs use each candidate's actual generated lengths in Eq.~\ref{eq:risk-cost}; the all-high reference provides a feasible fallback. We freeze selected actions, floors, and priorities in $\Lambda$, alongside reserve $\Gamma$ and budget $B_{\mathrm D}$.

\begin{algorithm}[t]
\caption{PASK Offline Calibration}
\label{alg:pask-offline}
\KwRequire{$\mathcal D_{\mathrm{cal}},\mathcal D_{\mathrm{dev}},\mathcal C,\mathcal T,\alpha$}
\KwEnsure{$\Lambda,\Gamma,B_{\mathrm D}$}
$(\bar V,D,\bar N)\gets\mathrm{Calibrate}(\mathcal D_{\mathrm{cal}},\mathcal C)$\;
$\rho_D\gets\mathrm{Priorities}(D,\bar N,m)$\;
$\mathcal P\gets\{\pi_{\mathrm{ref}}\}$\;
\ForAll{$\tau\in\mathcal T$}{
  $f\gets\mathrm{Floor}(\bar V;\tau)$\;
  \ForAll{$A\in\mathrm{GreedyPath}(f,\rho_D)$}{
    $B\gets\mathrm{ReferenceCostRatio}(A)$\;
    $\Gamma\gets\mathrm{Reserve}(\mathcal D_{\mathrm{cal}},f,B)$\;
    Add $\mathrm{Controller}(A,f,\rho_D,B,\Gamma)$ to $\mathcal P$\;
  }
}
Evaluate complete controllers on $\mathcal D_{\mathrm{dev}}$\;
\KwRet{$\mathrm{Pack}(\pi_{i^\star})$ using Eq.~\ref{eq:dev-selection}}\;
\end{algorithm}

\subsection{Structure-Conditioned Online Control}
\label{sec:online-lookup}
Each accepted non-stop token adds allowance $b=B_{\mathrm D}C_{\mathrm{full}}$. The balance $R_t$ includes the current allowance, with $R_1=b$. To anticipate upcoming protection costs, the controller looks up a reserve table constructed offline:
\begin{equation}
\Gamma(s;B_{\mathrm D})=Q_\beta\big(\{S^{(K)}_{r,u}(B_{\mathrm D}):s^{\mathrm{ref}}_{r,u}=s\}\big).
\label{eq:reserve}
\end{equation}
Here, $S^{(K)}_{r,u}$ is the largest positive cumulative floor cost minus newly accrued allowance over prefixes of the next $K$ reference tokens. We use $K=8$, $\beta=0.8$, and a pooled quantile for unseen states. Only the current parser state is needed online; Appendix~\ref{app:pask-budget-details} details the offline construction.

For the current structural tag, lookup returns base actions $a^0_{t,c}$, floors $a^-_{t,c}$, and frozen priorities for each Transformer layer group. PASK initializes
\begin{equation}
\begin{aligned}
a_{t,c}&=\max_{\preceq}(a^0_{t,c},a^-_{t,c}),\\
B_t&=[R_t-\Gamma(s_t;B_{\mathrm D})]_+.
\end{aligned}
\label{eq:online-budget}
\end{equation}
Writing $M(\mathbf a)=\sum_c m(a_c,c)$, it demotes eligible actions in ascending stored priority until $M(\mathbf a_t)\leq B_t$ or only floors remain. For critical or lookahead-critical tags, it promotes in descending priority only while total cost fits $B_t$. Floors take precedence over the target budget; negative balances carry forward. Adjustments affect only the current token's pending actions, never reconstructing previously discarded KV. Thus, the policy stays fixed while its current-token allocation adapts to accumulated budget.

Given resolved tags and constant-time lookup, control costs $O(|\mathcal C|\log|\mathcal C|)$ per token, independent of history length for fixed groups. This excludes parser adaptation, KV materialization, and attention computation. Appendix~\ref{app:pask-budget-details} specifies budget accounting and speculative-decoding rollback.

\subsection{Lightweight Prefill Companion}
\label{sec:prefill-companion}
A separate adapter groups prompt spans by semantic label and layer group. Table~\ref{tab:app-prefill-labels} lists protected content, function names, parameter keys, constraints, and schema scaffold, rather than output-parser transitions. Independently calibrated floors and distortion priorities determine a frozen retention plan before the prompt forward; its $R/L/H$ representations are materialized after high-precision prefill. Prefill and decode use separate tolerances and budgets; their composed accuracy and total KV cost are evaluated end to end. Appendix~\ref{app:prefill-decode-decomposition} reports stage-wise contributions.

\begin{algorithm}[t]
\caption{Online Structural Control}
\label{alg:pask-online}
\KwRequire{$z_t,s_t,R_t,B_{\mathrm D},\mathcal C,\Lambda,\Gamma$}
\KwEnsure{$\mathbf a_t,R_{t+1}$}
$(\mathbf a^0,\mathbf a^-,\boldsymbol\rho)\gets\mathrm{Lookup}(\Lambda,z_t,\mathcal C)$\;
$\mathbf a_t\gets\max_{\preceq}(\mathbf a^0,\mathbf a^-)$\;
$B_t\gets[R_t-\Gamma(s_t;B_{\mathrm D})]_+$\;
$\mathbf a_t\gets\mathrm{Demote}(\mathbf a_t,\mathbf a^-,B_t,\boldsymbol\rho)$\;
\If{$\mathrm{Crit}(z_t)$}{
  $\mathbf a_t\gets\mathrm{Promote}(\mathbf a_t,B_t,\boldsymbol\rho)$\;
}
Schedule $\mathbf a_t$ for KV materialization\;
$R_{t+1}\gets R_t-M(\mathbf a_t)+B_{\mathrm D}C_{\mathrm{full}}$\;
\KwRet{$\mathbf a_t,R_{t+1}$}\;
\end{algorithm}

\section{Experiments}
\label{sec:experiments}
We directly test the central claim: whether parser-derived structural state can sustain structured-generation correctness at substantially reduced KV cost. Evaluation covers task accuracy, serving efficiency, ablations, and sensitivity analyses.

\begin{table*}[t]
\centering
\footnotesize
\setlength{\fboxsep}{0.5pt}
\setlength{\tabcolsep}{2pt}
\renewcommand{\arraystretch}{0.85}
\resizebox{\textwidth}{!}{%
\begin{tabular}{@{}c|c|ccccc|ccccc@{}}
\toprule
\multirow[c]{2}{*}[-0.5ex]{\textbf{Model}} & \multirow[c]{2}{*}[-0.5ex]{\textbf{Method}} & \multicolumn{5}{c|}{\textbf{BFCL non-live}} & \multicolumn{5}{c}{\textbf{BFCL live}} \\
\cmidrule(lr){3-7}\cmidrule(lr){8-12}
& & \textbf{Simple} & \textbf{Multiple} & \textbf{Parallel} & \textbf{Par.-multiple} & \textbf{Overall} & \textbf{Simple} & \textbf{Multiple} & \textbf{Parallel} & \textbf{Par.-multiple} & \textbf{Overall} \\
\midrule
\multirow[c]{10}{*}{\rotatebox[origin=c]{90}{\textbf{Qwen3-4B}}}
& \textcolor{gray}{Full KV} & \textcolor{gray}{95.00} & \textcolor{gray}{95.00} & \textcolor{gray}{90.00} & \textcolor{gray}{84.17} & \textcolor{gray}{91.83} & \textcolor{gray}{78.57} & \textcolor{gray}{74.57} & \textcolor{gray}{60.00} & \textcolor{gray}{78.57} & \textcolor{gray}{75.22} \\
& H2O \venuebox{NeurIPS 2023} & 75.83 & 81.67 & 31.67 & 19.17 & 56.83 & 45.45 & 48.18 & 0.00 & 21.43 & 46.61 \\
& SnapKV\venuebox{NeurIPS 2024} & 85.00 & 87.50 & 42.50 & 35.00 & 67.00 & 55.19 & 59.08 & 0.00 & 14.29 & 56.84 \\
& MiKV\venuebox{arXiv 2024} & 87.50 & 90.83 & \resultsecond{76.67} & \resultsecond{76.67} & \resultsecond{83.83} & \resultsecond{64.94} & 64.45 & 30.00 & \resultbest{50.00} & \resultsecond{63.87} \\
& RocketKV \venuebox{ICML 2025} & 75.42 & 65.83 & 33.33 & 25.00 & 55.00 & 48.05 & 47.24 & 0.00 & \resultsecond{28.57} & 46.49 \\
& CriticalKV\venuebox{ICML 2026} & 87.92 & \resultsecond{91.67} & 59.17 & 40.00 & 73.33 & 59.74 & \resultsecond{66.19} & 30.00 & \resultsecond{28.57} & \resultsecond{63.87} \\
& LazyEviction\venuebox{ACL 2026} & 46.67 & 81.67 & 20.83 & 14.17 & 42.00 & 33.12 & 60.03 & 10.00 & \resultsecond{28.57} & 53.76 \\
& TriAxialKV\venuebox{arXiv 2026} & \resultsecond{90.42} & 85.00 & 70.83 & 56.67 & 78.67 & 62.99 & 59.08 & \resultsecond{40.00} & 0.00 & 58.57 \\
& TransAttn\venuebox{arXiv 2026} & 88.75 & 64.17 & 54.17 & 43.33 & 67.83 & 45.45 & 27.49 & 0.00 & 0.00 & 30.09 \\
& \textbf{PASK (Ours)} & \resultbest{92.08} & \resultbest{93.33} & \resultbest{80.83} & \resultbest{81.67} & \resultbest{88.00} & \resultbest{74.03} & \resultbest{72.20} & \resultbest{60.00} & \resultbest{50.00} & \resultbest{72.01} \\
\midrule
\multirow[c]{10}{*}{\rotatebox[origin=c]{90}{\textbf{LLaMA-3.1-8B}}}
& \textcolor{gray}{Full KV} & \textcolor{gray}{94.58} & \textcolor{gray}{95.00} & \textcolor{gray}{91.67} & \textcolor{gray}{80.83} & \textcolor{gray}{91.33} & \textcolor{gray}{75.32} & \textcolor{gray}{60.98} & \textcolor{gray}{40.00} & \textcolor{gray}{57.14} & \textcolor{gray}{63.38} \\
& H2O\venuebox{NeurIPS 2023} & 75.42 & 90.00 & 15.83 & 12.50 & 53.83 & 46.75 & 36.65 & 0.00 & 14.29 & 37.73 \\
& SnapKV\venuebox{NeurIPS 2024} & 87.92 & 94.17 & 43.33 & 35.00 & 69.67 & 51.30 & \resultsecond{53.55} & 10.00 & \resultsecond{35.71} & 52.28 \\
& MiKV\venuebox{arXiv 2024} & 91.67 & \resultsecond{95.00} & 66.67 & 51.67 & 79.33 & 64.94 & 48.50 & 20.00 & 28.57 & 50.92 \\
& RocketKV\venuebox{ICML 2025} & 82.92 & 90.83 & 25.00 & 21.67 & 60.67 & 51.30 & 47.39 & 0.00 & 21.43 & 47.10 \\
& CriticalKV\venuebox{ICML 2026} & 86.67 & 93.33 & 35.83 & 33.33 & 67.17 & 62.34 & 53.08 & 10.00 & \resultsecond{35.71} & \resultsecond{54.01} \\
& LazyEviction\venuebox{ACL 2026} & 56.67 & 88.33 & 25.83 & 16.67 & 48.83 & 38.96 & 45.66 & 10.00 & 28.57 & 43.65 \\
& TriAxialKV\venuebox{arXiv 2026} & \resultsecond{92.08} & 88.33 & \resultsecond{85.83} & \resultsecond{55.83} & \resultsecond{82.83} & \resultsecond{66.23} & 38.23 & \resultbest{50.00} & 28.57 & 43.53 \\
& TransAttn\venuebox{arXiv 2026} & 90.00 & 67.50 & 71.67 & 38.33 & 71.50 & 60.39 & 32.07 & 30.00 & 0.00 & 36.87 \\
& \textbf{PASK (Ours)} & \resultbest{93.75} & \resultbest{96.67} & \resultbest{90.83} & \resultbest{80.83} & \resultbest{91.17} & \resultbest{73.38} & \resultbest{58.45} & \resultsecond{40.00} & \resultbest{50.00} & \resultbest{60.91} \\
\midrule
\multirow[c]{10}{*}{\rotatebox[origin=c]{90}{\textbf{Qwen3-14B}}}
& \textcolor{gray}{Full KV} & \textcolor{gray}{96.25} & \textcolor{gray}{98.33} & \textcolor{gray}{92.50} & \textcolor{gray}{81.67} & \textcolor{gray}{93.00} & \textcolor{gray}{79.87} & \textcolor{gray}{75.20} & \textcolor{gray}{70.00} & \textcolor{gray}{64.29} & \textcolor{gray}{75.83} \\
& H2O\venuebox{NeurIPS 2023} & 72.08 & 85.00 & 22.50 & 16.67 & 53.67 & 54.55 & 51.82 & 0.00 & 0.00 & 50.80 \\
& SnapKV\venuebox{NeurIPS 2024} & 82.50 & \resultsecond{91.67} & 60.00 & 55.83 & 74.50 & 61.69 & \resultsecond{66.98} & 0.00 & 14.29 & 64.24 \\
& MiKV\venuebox{arXiv 2024} & 83.75 & 88.33 & \resultsecond{79.17} & \resultsecond{65.83} & \resultsecond{80.17} & 66.23 & 63.03 & \resultbest{40.00} & \resultsecond{42.86} & 63.01 \\
& RocketKV\venuebox{ICML 2025} & 78.33 & 90.00 & 30.00 & 41.67 & 63.67 & 55.19 & 55.13 & \resultsecond{30.00} & \resultsecond{42.86} & 54.62 \\
& CriticalKV\venuebox{ICML 2026} & 80.42 & 87.50 & 54.17 & 50.83 & 70.67 & 61.69 & 65.72 & \resultbest{40.00} & \resultbest{57.14} & \resultsecond{64.49} \\
& LazyEviction\venuebox{ACL 2026} & 49.17 & \resultsecond{91.67} & 28.33 & 25.00 & 48.67 & 33.77 & 64.14 & 20.00 & \resultbest{57.14} & 57.71 \\
& TriAxialKV\venuebox{arXiv 2026} & 87.92 & 79.17 & 77.50 & 64.17 & 79.33 & \resultsecond{68.18} & 61.14 & \resultsecond{30.00} & 0.00 & 61.04 \\
& TransAttn\venuebox{arXiv 2026} & \resultsecond{88.33} & 83.33 & 70.00 & 49.17 & 75.83 & 49.35 & 31.12 & 0.00 & 0.00 & 33.66 \\
& \textbf{PASK (Ours)} & \resultbest{92.08} & \resultbest{94.17} & \resultbest{86.67} & \resultbest{79.17} & \resultbest{88.83} & \resultbest{73.38} & \resultbest{73.62} & \resultsecond{30.00} & \resultbest{57.14} & \resultbest{72.75} \\
\bottomrule
\end{tabular}%
}
\vspace{-0.1in}
\caption{BFCL accuracy (\%) across models and the non-live and Live subsets. Red and orange shading indicate the \colorbox{red!10}{\textbf{best}} and \colorbox{orange!15}{\underline{second-best}} accuracy, respectively; Full KV is shown as a reference and excluded from ranking.}
\label{tab:main-results}
\vspace{-0.1in}
\end{table*}

\subsection{Experimental Setup}
\label{sec:exp-setup}

\textbf{Models and Datasets.}
We evaluate PASK with Qwen3-4B-Instruct-2507, Qwen3-14B~\citep{yang2025qwen3technicalreport}, and LLaMA-3.1-8B-Instruct~\citep{grattafiori2024llama3herdmodels}, abbreviated as Qwen3-4B, Qwen3-14B, and LLaMA-3.1-8B, respectively. Our benchmarks cover BFCL function calling constrained by tool specifications~\citep{pmlr-v267-patil25a}, including both non-live and Live subsets, and BIRD text-to-SQL generation of executable, schema-grounded SQLite queries~\citep{NEURIPS2023_83fc8fab}. BIRD experiments use Qwen3-4B and the official Dev package; Table~\ref{tab:app-datasets} reports dataset statistics.

\textbf{Baselines.}
We compare against eight baselines covering attention-based retention, mixed-precision storage, recurrence-aware eviction, and structured cache policies: H2O~\citep{NEURIPS2023_6ceefa7b}, SnapKV~\citep{NEURIPS2024_28ab4182}, MiKV~\citep{yang2024tokenleftbehindreliable}, RocketKV~\citep{pmlr-v267-behnam25a}, CriticalKV~\citep{pmlr-v306-feng26j}, LazyEviction~\citep{zhang-etal-2026-lazyeviction}, TriAxialKV~\citep{shen2026triaxialkvextremelowprecisionkvcache}, and Transactional Attention~\citep{basu2026transactionalattentionsemanticsponsorship}. All share prompts, decoding configurations, and the constrained-decoding backend~\citep{MLSYS2025_5c20ca4b}.

\textbf{Evaluation \& Implementation.}
Calibration, selection, and evaluation use disjoint calibration, development, and frozen test splits partitioned by normalized tool-call schemas (BFCL) or databases (BIRD). Appendix~\ref{app:baseline-tuning} details baseline-specific tuning. Qwen3-4B serving on NVIDIA L40S and A100 uses matched software and decoding settings. $B$ denotes measured, workload-weighted total KV cost (prefill and decode), normalized by measured Full KV ($B=1.00$). PASK achieves $B\approx0.33$ in the main comparison; flexible-budget baselines are configured to match this cost, while fixed-profile methods retain native configurations with measured costs reported separately.
\subsection{Main Results}
\label{sec:exp-main}

\begin{figure*}[t]
    \centering
    \includegraphics[width=\textwidth, trim=6mm 3mm 13mm 3mm, clip]{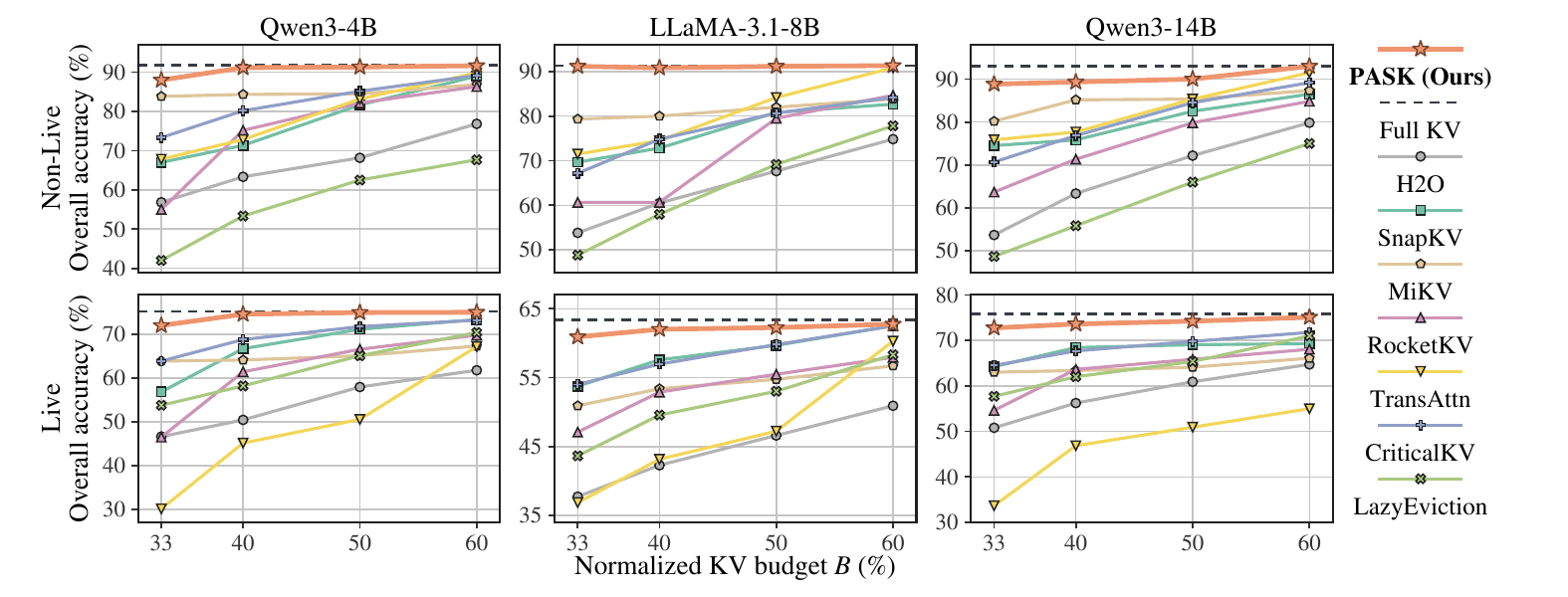}
    \vspace{-0.3in}
    \caption{BFCL Overall accuracy versus measured relative total KV cost $B$ across three models and both subsets.}
    \label{fig:bfcl-budget-grid}
    \vspace{-0.1in}
\end{figure*}

\begin{figure*}[t]
\centering
\includegraphics[width=\textwidth, trim=0mm 0mm 2mm 0mm, clip]{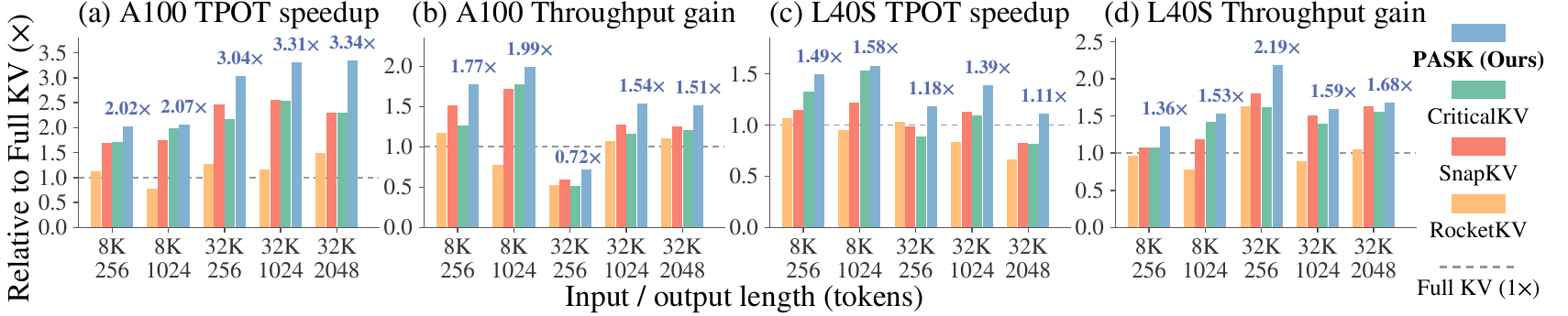}
\vspace{-0.3in}
\caption{Relative serving efficiency for Qwen3-4B at $c=8$ and approximate total KV budget $B=0.33$. (a,c) TPOT speedup, defined as $\mathrm{TPOT}_{\mathrm{Full}}/\mathrm{TPOT}_{\mathrm{method}}$. (b,d) Throughput gain, defined as $\mathrm{TPS}_{\mathrm{method}}/\mathrm{TPS}_{\mathrm{Full}}$.}
\label{fig:system-efficiency}
\vspace{-0.1in}
\end{figure*}

\textbf{Task accuracy.}
\circnum{1} \textbf{Consistent accuracy gains.} PASK leads all evaluated compressed methods across six BFCL Overall settings in Table~\ref{tab:main-results} and all four BIRD budgets in Table~\ref{tab:bird-results}. On Live BFCL, Qwen3-4B and Qwen3-14B improve over the strongest compressed baselines by $8.14$ and $8.26$ points, respectively. Fig.~\ref{fig:bfcl-budget-grid} shows its BFCL lead across budgets.  \circnum{2} \textbf{Near-Full-KV compositional accuracy.} On LLaMA-3.1-8B non-Live, PASK nearly matches Full KV in Overall accuracy ($91.17\%$ vs.\ $91.33\%$) and matches it on parallel-multiple calls ($80.83\%$), outperforming the strongest compressed baseline by $25.0$ points. Across all three models, parallel-multiple accuracy remains within $2.5$ points of Full KV. \circnum{3} \textbf{Accuracy under tighter KV budgets.} On BIRD, PASK achieves $52.71\%$ execution accuracy at $B=0.275$, exceeding the strongest same-budget baseline ($49.48\%$) and even the best baseline at $B=0.40$ ($51.83\%$). At $B=0.60$, it reaches $55.93\%$, only $0.35$ points below Full KV. Detailed BFCL budget results appear in Appendix~\ref{app:multibudget-accuracy}.

\begin{table}[t]
\centering
\footnotesize
\setlength{\fboxsep}{0.5pt}
\setlength{\tabcolsep}{3pt}
\renewcommand{\arraystretch}{0.9}
\resizebox{\linewidth}{!}{%
\begin{tabular}{@{}lccccc@{}}
\toprule
\multirow[c]{2}{*}[-0.4em]{\textbf{Method}} & \multicolumn{5}{c}{\textbf{KV budget $B$}} \\
\cmidrule(lr){2-6}
& \textcolor{gray}{\textbf{100\%}} & \textbf{60\%} & \textbf{50\%} & \textbf{40\%} & \textbf{27.5\%} \\
\midrule
H2O & \textcolor{gray}{56.28} & 41.10 & 35.51 & 32.55 & 24.08 \\
SnapKV & \textcolor{gray}{56.28} & 49.83 & 46.77 & 44.24 & 36.56 \\
MiKV & \textcolor{gray}{56.28} & 54.01 & 52.27 & 50.26 & 48.25 \\
RocketKV & \textcolor{gray}{56.28} & 54.36 & \resultsecond{53.66} & 50.09 & 41.10 \\
CriticalKV & \textcolor{gray}{56.28} & 50.96 & 44.68 & 41.80 & 37.52 \\
LazyEviction & \textcolor{gray}{56.28} & \resultsecond{55.67} & 52.88 & 50.96 & \resultsecond{49.48} \\
TransAttn & \textcolor{gray}{56.28} & 54.80 & 53.05 & \resultsecond{51.83} & 40.14 \\
\textbf{PASK (Ours)} & \textcolor{gray}{56.28} & \resultbest{55.93} & \resultbest{53.84} & \resultbest{53.75} & \resultbest{52.71} \\
\bottomrule
\end{tabular}%
}
\vspace{-0.1in}
\caption{Execution accuracy (EX, \%) on BIRD with Qwen3-4B. Gray entries show the Full KV reference.}
\label{tab:bird-results}
\vspace{-0.1in}
\end{table}

\textbf{Memory.}
At a fixed logical capacity of 229,896 tokens, PASK achieves the lowest peak GPU memory across both evaluated GPUs in Table~\ref{tab:peak-memory}, reducing Full-KV memory by 47.5\% and using 13.8\% less than RocketKV, the strongest compressed comparator in memory.

\begin{table}[t]
\centering
\footnotesize
\setlength{\tabcolsep}{2pt}
\renewcommand{\arraystretch}{0.9}
\resizebox{\linewidth}{!}{%
\begin{tabular}{@{}c|cc|cc@{}}
\toprule
\multirow[c]{2}{*}[-0.4em]{\textbf{Method}} & \multicolumn{2}{c|}{\textbf{NVIDIA L40S}} & \multicolumn{2}{c}{\textbf{NVIDIA A100}} \\
\cmidrule(lr){2-3}\cmidrule(lr){4-5}
& \textbf{Peak (MiB)} $\downarrow$ & \textbf{Saving} $\uparrow$ & \textbf{Peak (MiB)} $\downarrow$ & \textbf{Saving} $\uparrow$ \\
\midrule
Full KV & 40986 & --- & 41008 & --- \\
SnapKV & 25090 & 38.78\% & 25086 & 38.83\% \\
RocketKV & 24976 & 39.06\% & 24976 & 39.09\% \\
CriticalKV & 24976 & 39.06\% & 24978 & 39.09\%  \\
\rowcolor{oursrow}
\textbf{PASK (Ours)} & \textbf{21528} & \textbf{47.47\%} & \textbf{21536} & \textbf{47.48\%} \\
\bottomrule
\end{tabular}}
\vspace{-0.1in}
\caption{Peak GPU memory for Qwen3-4B at a fixed logical capacity of 229,896 tokens.}
\vspace{-0.1in}
\label{tab:peak-memory}
\end{table}

\textbf{Speed.}
PASK achieves the lowest TPOT and highest throughput among compressed baselines across all ten GPU/workload settings in Fig.~\ref{fig:system-efficiency}, reaching 3.34$\times$ TPOT speedup on A100 (32K/2048) and 2.19$\times$ throughput on L40S (32K/256) relative to Full KV. Against the strongest compressed comparator per setting, it reduces TPOT by up to 31.1\% and improves throughput by up to 25.9\%. At the A100 32K/256 setting, PASK's throughput remains below that of Full KV despite a 3.04$\times$ TPOT speedup. This is consistent with packed INT4 storage reducing KV traffic and frozen parser-conditioned lookup avoiding online importance estimation.

\subsection{Ablation Study}
\label{sec:exp-ablation}

\begin{figure}[t]
\centering
\includegraphics[width=1.0\linewidth, trim=0mm 1mm 4mm 0mm, clip]{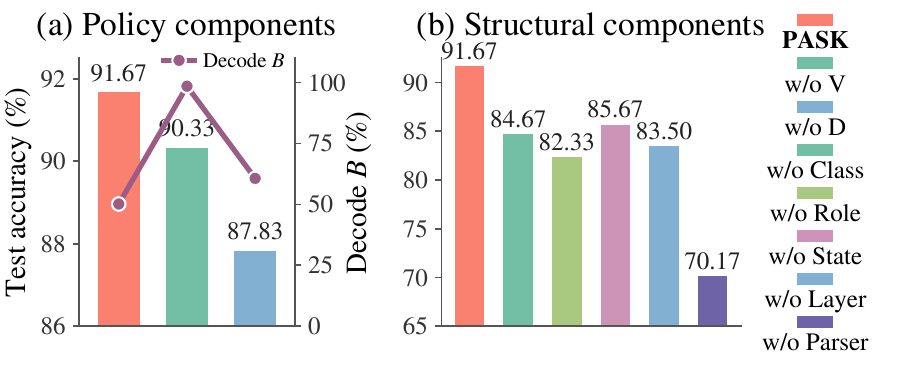}
\vspace{-0.4in}
\caption{PASK ablations. (a) Policy components: accuracy (bars) and Decode budget (line). (b) Structural components: accuracy at an aligned Decode budget.}
\label{fig:ablation-summary}
\vspace{-0.10in}
\end{figure}

\begin{figure*}[t]
\centering
\includegraphics[width=\linewidth]{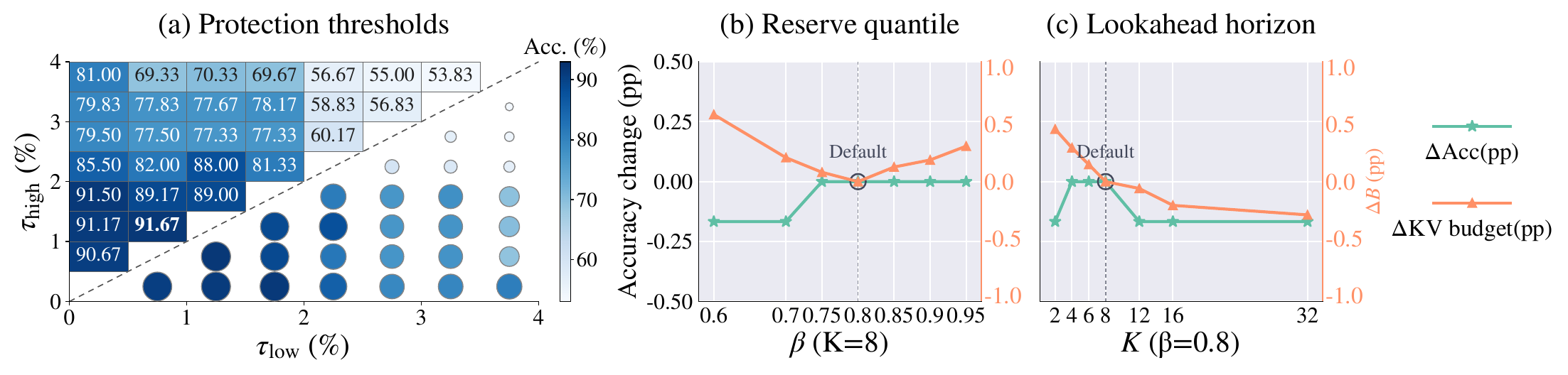}
\vspace{-0.4in}
\caption{Hyperparameter sensitivity on Qwen3-4B BFCL non-Live. (a) Accuracy across 28 threshold pairs; bubbles mirror the heatmap, and bold marks the selected pair. (b,c) Accuracy and Decode-budget changes (pp) for $\beta$ and $K$, relative to the default $(0.8,8)$.}
\label{fig:tau-sensitivity-heatmap}
\vspace{-0.1in}
\end{figure*}

\circnum{1} \textbf{Parser-derived signals improve retention.}
In Fig.~\ref{fig:ablation-summary}, at matched Decode budgets, removing structural roles, layer grouping, token classes, or parser state and lookahead reduces accuracy by $9.33$, $8.17$, $7.00$, and $6.00$ points, respectively. Removing parser-transition information entirely causes a larger $21.50$-point drop, supporting its importance for KV retention. Appendix~\ref{app:experimental-details} details ablation protocols. \circnum{2} \textbf{Task protection reduces budgets.} Removing V-based protection floors increases Decode budget from $50.15\%$ to $98.49\%$ while lowering accuracy from $91.67\%$ to $90.33\%$ in Fig.~\ref{fig:ablation-summary}. Thus, distortion ranking alone requires substantially more memory to meet the development reliability criterion. \circnum{3} \textbf{Distortion ranking improves residual allocation.} Keeping protection floors, replacing D-ranking with fixed-order, layer-group round-robin allocation increases Decode budget by $10.52$ points and reduces accuracy by $3.83$ points in Fig.~\ref{fig:ablation-summary}, confirming that distortion ranking complements task-error sensitivity in residual allocation.

\subsection{Hyperparameter Analysis}
\label{sec:exp-hyperparameter}

Fig.~\ref{fig:tau-sensitivity-heatmap}(a) evaluates 28 protection-threshold pairs for Qwen3-4B on BFCL non-Live at approximately 50\% decode KV budget. Pairs with $\tau_{\mathrm{high}}\leq1\%$ achieve $90.67$--$91.67\%$ accuracy, versus $53.83$--$60.17\%$ for $\tau_{\mathrm{low}}\geq2\%$, highlighting the cost of relaxed protection. The selected $(0.5\%,1\%)$ pair achieves $91.67\%$ accuracy at realized decode budget $B_{\mathrm D}=0.5015$.

The reserve quantile $\beta$ and lookahead horizon $K$ control budget reservation for upcoming protected tokens. Fig~\ref{fig:tau-sensitivity-heatmap}(b,c) varies one parameter while fixing the other at its corresponding default value and reports accuracy and realized Decode KV-budget changes relative to $\beta=0.8$ and $K=8$. Stage-specific reliability tolerances $\alpha_{\mathrm D}$ and $\alpha_{\mathrm P}$ govern development-set policy selection; their quality--cost trade-offs are reported in Appendix~\ref{app:alpha-selection}.
\subsection{Additional Analysis}
\label{sec:exp-transfer}

\textbf{Responsibility--distortion mismatch.}
Table~\ref{tab:mismatch-correlation} in Appendix~\ref{app:mismatch-evidence} provides evidence for separating task protection from numerical fidelity. Across 27 Qwen3-4B BFCL non-Live calibration buckets, task-error sensitivity and distortion yield disjoint top-20\% sets for both release and low-precision retention. Of the most task-sensitive buckets, 100\% and 83.3\% fall in the respective bottom halves of distortion rankings. Thus, low numerical distortion need not imply low task risk.

\textbf{Calibration cost and reuse.}
Table~\ref{tab:calibration-cost} reports about 3.3 GPU-hours for combined Runtime-V/D calibration on one A100-80GB (Qwen3-4B, 200 BFCL non-Live cases per candidate). Table~\ref{tab:calibration-transfer} compares unchanged transfer (trans) and per-target recalibration (recalib) at $B\in\{0.33,0.50\}$; Best baseline denotes each setting's strongest compressed comparator. Transfer outperforms this baseline across all three targets and both budgets, even without target-specific policy recalibration. On Qwen3-14B non-Live, it differs from recalibration by at most 0.83 points, outperforming it by 0.83 at $B=0.50$. Recalibration gains on LLaMA-3.1-8B non-Live and Qwen3-4B Live fall from 3.84/2.34 points at $B=0.33$ to 1.67/0.99 at $B=0.50$, respectively, showing effective transfer across evaluated targets and greater recalibration benefits under tighter KV budgets.

\textbf{High-concurrency scaling.}
At concurrency 32 on A100 (32K inputs, 1024-token checkpoint), PASK achieves $2.17\times$ throughput and $4.84\times$ TPOT speedups over Full KV in Table~\ref{tab:app-system-scaling}, while maintaining 68--69 ms TPOT across concurrency 8,16,32 as throughput rises from 51.02 to 72.23 tokens/s, demonstrating scalability under increasing concurrency in long-context serving.

\begin{table}[t]
\centering
\footnotesize
\setlength{\tabcolsep}{3pt}
\renewcommand{\arraystretch}{1}
\resizebox{\linewidth}{!}{%
\begin{tabular}{@{}clccc@{}}
\toprule
\multirow[c]{2}{*}{$B$} & \multirow[c]{2}{*}{\textbf{Method}} & \textbf{Qwen3-4B} & \textbf{Qwen3-14B} & \textbf{LLaMA-3.1-8B} \\
& & Live & non-Live & non-Live \\
\midrule
\multirow[c]{3}{*}{0.33} & Best baseline & 63.87 & 80.17 & 82.83 \\
& PASK (trans) & 69.67 & 88.67 & 87.33 \\
& PASK (recalib) & 72.01 & 88.83 & 91.17 \\
\midrule
\multirow[c]{3}{*}{0.50} & Best baseline & 71.76 & 85.33 & 84.17 \\
& PASK (trans) & 73.98 & 90.83 & 89.50 \\
& PASK (recalib) & 74.97 & 90.00 & 91.17 \\
\bottomrule
\end{tabular}%
}
\vspace{-0.1in}
\caption{BFCL accuracy (\%) under calibration transfer. 
}

\label{tab:calibration-transfer}
\vspace{-0.05in}
\end{table}

\section{Conclusion}
PASK connects output legality with memory retention by reusing parser transitions already computed during constrained decoding. Calibrated against task errors, this structural signal guides KV protection across Transformer layers, separating task protection from numerical fidelity. Function-calling and text-to-SQL experiments demonstrate stronger accuracy under constrained budgets, complemented by serving efficiency and policy reuse. Future work will explore adaptive recalibration and agents alternating between free-form reasoning and constrained outputs. More broadly, PASK illustrates a principle for efficient generation: reuse available structural information to allocate memory according to the consequences of losing history, aligning compression with task correctness.

\section*{Limitations}

PASK relies on offline calibration to construct a structure-conditioned policy table. In the current design, this calibration is performed per model and grammar family before deployment; transfer across model families and to the Live split concedes one to four points to per-target recalibration on some targets (Section~\ref{sec:exp-transfer}). The implementation also discretizes depth into three layer groups and persistence into three actions; finer partitions or additional precision levels may offer greater flexibility but are not evaluated here. Finally, the method requires a formal constraint engine that exposes parser state during decoding; extending PASK to requests that alternate free-form language with structured calls is future work discussed in Appendix~\ref{app:sample-compression-cases}.




\bibliography{custom}
\clearpage
\appendix

\section{Structural Bucket Construction}
\label{app:policy-construction}
\providecommand{\appyes}{\begingroup\setlength{\fboxsep}{1.2pt}\colorbox{oursrow}{\ensuremath{\checkmark}}\endgroup}
\providecommand{\appno}{\textcolor{black!45}{\ensuremath{\times}}}
A deterministic adapter converts grammar information already computed for constrained decoding into lookup keys; offline task feedback determines their protection.

\subsection{Decode tags and structural attribution}
Table~\ref{tab:app-tag-schema} lists the task-specific vocabularies for the structural tags defined in Eq.~\ref{eq:structural-bucket}. BFCL assigns roles according to function dispatch and required or optional arguments, whereas BIRD uses SQL productions and schema context.

Ambiguous transitions and tokens spanning structural boundaries retain candidate tags for conservative action and floor resolution (Appendix~\ref{app:backend}). Unseen schemas use context and parent fallback; supporting a new grammar family requires an adapter implementing the same four-field interface.

\begin{table*}[t]
\centering
\scriptsize
\setlength{\tabcolsep}{4pt}
\renewcommand{\arraystretch}{1.15}
\begin{tabular}{@{}l p{0.40\textwidth} p{0.40\textwidth}@{}}
\toprule
\textbf{Component} & \textbf{Function calling (BFCL)} & \textbf{Text-to-SQL (BIRD)} \\
\midrule
Class & \texttt{pure\_scaffold}, \texttt{schema\_bearing}, \texttt{free\_generation} & \texttt{pure\_scaffold}, \texttt{schema\_bearing}, \texttt{free\_generation} \\
Role & \texttt{scaffold\_separator}, \texttt{required}, \texttt{optional}, \texttt{dispatch\_key} & \texttt{projection}, \texttt{relation}, \texttt{predicate}, \texttt{outside\_schema} \\
State & \texttt{boundary}, \texttt{critical} & \texttt{stable}, \texttt{boundary}, \texttt{critical} \\
Next & \texttt{lookahead\_critical}, \texttt{no\_lookahead} & \texttt{lookahead\_critical}, \texttt{no\_lookahead} \\
\bottomrule
\end{tabular}
\vspace{-0.1in}
\caption{Task-specific vocabularies for the four structural-tag components.}
\label{tab:app-tag-schema}
\end{table*}
\subsection{Layer groups and lookup keys}
For $L$ layers, let $b_1=\lfloor(L+2)/3\rfloor$ and $b_2=\lfloor(2L+2)/3\rfloor$. The lower, middle, and upper groups are $[0,b_1)$, $[b_1,b_2)$, and $[b_2,L)$, respectively. Their zero-indexed ranges are 0--11/12--23/24--35 for Qwen3-4B, 0--13/14--26/27--39 for Qwen3-14B, and 0--10/11--21/22--31 for LLaMA-3.1-8B. Payload costs use the actual dimensions of each group.

Decode and prefill lookup keys are $q^{\mathrm D}_{t,c}=(z_t,c)$ and $q^{\mathrm P}_{t,c}=(\mathrm{label}_t,c)$, respectively. Each calibrated entry stores its base action, protection floor, and adjacent-upgrade scores.

\subsection{Prompt-span labels}
Prompt labels come from deterministic span matching. All text starts as \texttt{protected\_content}. Validated BFCL tool-schema JSON and BIRD SQLite DDL spans become \texttt{schema\_scaffold}, with finer labels assigned by serialization-specific patterns (Table~\ref{tab:app-prefill-labels}); invalid or unrecognized spans retain the default. No human annotation or general-purpose AST parser is used; new serializations require new patterns.

Span offsets map labels to tokens. Cross-boundary tokens conservatively combine the corresponding actions and floors (Appendix~\ref{app:backend}), without imposing a universal risk order on labels. \texttt{protected\_content} is a category, not necessarily an $H$ action; sink and recent-window masks independently enforce high retention.

\begin{table*}[t]
\centering
\footnotesize
\setlength{\tabcolsep}{4pt}
\renewcommand{\arraystretch}{1.15}
\begin{tabular}{@{}l p{0.75\textwidth}@{}}
\toprule
\textbf{Semantic label} & \textbf{Assignment rule} \\
\midrule
Protected content & Prompt text outside validated schema spans: instructions, requests, and chat context. \\
Schema scaffold & JSON/DDL structure and unmatched tokens within validated schema spans. \\
Function name & BFCL tool/function names; BIRD table identifiers. \\
Parameter key & BFCL property names and \texttt{required} entries; BIRD column identifiers. \\
Type constraint & BFCL \texttt{type} values; BIRD declared column types. \\
Enum constraint & BFCL \texttt{enum} values; recognized BIRD constraints, e.g., \texttt{PRIMARY KEY}, \texttt{REFERENCES}, and \texttt{NOT NULL}. \\
Schema description & BFCL \texttt{description} text. \\
\bottomrule
\end{tabular}
\vspace{-0.1in}
\caption{Semantic labels for prefill buckets in BFCL and BIRD.}
\label{tab:app-prefill-labels}
\vspace{-0.1in}
\end{table*}

\subsection{Calibration and frozen-policy records}
\label{app:pask-calibration-details}
Calibration follows Section~\ref{sec:offline-calibration} and Algorithm~\ref{alg:pask-offline}. Unobserved buckets use fallback mappings. Affected reference query positions are counted once per bucket and shared across distortion actions.

Within each threshold pair, adjacent upgrades use a fixed serialized bucket/action tie order; zero-score upgrades remain eligible until the all-high endpoint. Candidates retain their floor maps, so identical action maps with different floors remain distinct controllers. Nominal reference-trajectory cost is nondecreasing within each upgrade path, but autoregressive accuracy and realized cost need not be. The all-high fallback controller uses $B_{\mathrm D}=1$, all-high floors, and $\Gamma=0$.

The selected frozen policy records adapter mappings, tag vocabularies, layer boundaries, quantizer and payload-cost conventions, sink and recent-window settings, thresholds, stage budgets, reserve horizon and quantile, fallback maps, and tie order. These records identify the configuration used for evaluation and policy transfer.

\section{KV-Cache Backend}
\label{app:backend}
\begin{figure*}[t]
\centering
\includegraphics[width=\textwidth]{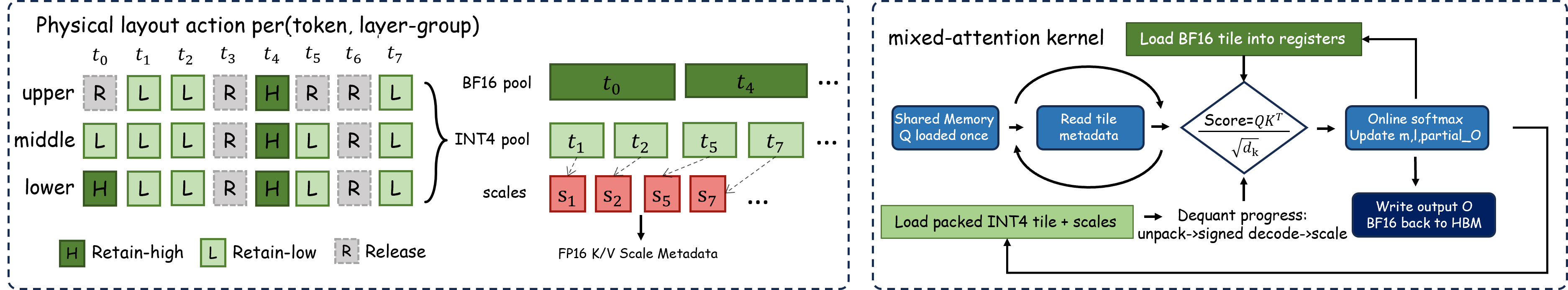}
\vspace{-0.25in}
\caption{PASK backend. Token--layer KV is retained in BF16/INT4 or released after recent protection. Disjoint index streams, on-demand dequantization, and shared FP32 online softmax support mixed-precision attention.}
\label{fig:pask-kernel-schematic}
\vspace{-0.1in}
\end{figure*}

Fig.~\ref{fig:pask-kernel-schematic} summarizes parser-conditioned storage and attention. Scheduled cost, live KV payload, and physical GPU memory are accounted for separately.

\subsection{Parser hand-off and resolution}
SGLang~\citep{NEURIPS2024_724be447} receives XGrammar~\citep{MLSYS2025_5c20ca4b} frontiers before and after each accepted non-stop token. Each frontier is a sorted state set with fields \textit{rule\_id}, \textit{sequence\_id}, \textit{element\_id}, and \textit{sub\_element\_id}. The serialized $F_{t-1}\to F_t$, schema key, and context resolve $z_t$ before the next forward pass produces the token's KV. Lookup uses $z_t\mathbin{::}\{\mathrm{lower},\mathrm{middle},\mathrm{upper}\}$. Missing or ambiguous mappings use schema context, then
\[
\begin{aligned}
(\mathrm{class},\mathrm{role},\mathrm{state},\mathrm{next})
&\to(\mathrm{class},\mathrm{role},\mathrm{state})\\
&\hspace{-1.5em}\to(\mathrm{class},\mathrm{role})\to\mathrm{class}.
\end{aligned}
\]
For candidate descendants $\mathcal U$, group $c$ receives
\begin{equation}
\widetilde a^0_c=\widetilde a^-_c=\max_{\preceq}\{a^0_{u,c},a^-_{u,c}:u\in\mathcal U\}.
\label{eq:app-fallback}
\end{equation}
Thus budget adjustment cannot undo conservative resolution. Optional promotions require a unique calibrated score; no compatible entry defaults to $H$ for both action and floor. Cross-label prompt tokens follow the same rule. Parent aggregates are precomputed; frontier adaptation is separate from constant-time lookup. Local conservatism does not guarantee unseen-schema accuracy.

\subsection{Mixed-precision pools and protection}
Layers share group-level actions but maintain separate KV tensors and slot maps. Current KV remains available during its forward pass, and sink/recent-window masks enforce $H$ until protection expires. Scheduled $L/R$ actions then trigger quantization or release. Subsequent promotion cannot reconstruct discarded or pre-quantization values. Slot maps preserve absolute positions across released entries.

\subsection{Payload cost and physical accounting}
\label{app:pask-accounting}
For KV-head count $H^{\mathrm{KV}}_\ell$ and head dimension $d_\ell$, group-wise K/V element and scale counts are
\begin{equation}
n_c=2\sum_{\ell\in c}H^{\mathrm{KV}}_\ell d_\ell,
\quad k_c=2\sum_{\ell\in c}H^{\mathrm{KV}}_\ell.
\end{equation}
For even dimensions, scheduled bit costs are
\begin{equation}
\begin{aligned}
m(R,c)&=0,\qquad m(H,c)=16n_c,\\
m(L,c)&=4n_c+16k_c.
\end{aligned}
\label{eq:action-cost}
\end{equation}
Odd dimensions add $4p_c$ padding bits to $m(L,c)$, where $p_c=2\sum_{\ell\in c}H^{\mathrm{KV}}_\ell(d_\ell\bmod2)$. Symmetric quantization needs no zero points. Group-specific costs give $C_{\mathrm{full}}=\sum_c m(H,c)$.

The ledger charges long-range schedules, so $R$ may still occupy protected-window storage. Live payload counts prompt/generated KV, protection, and scales without duplicating aliases. Physical memory additionally includes indices, alignment, pages, allocator overhead, and transient buffers. Peak GPU memory is measured across prefill, migration, and decode, not inferred from $C_{\mathrm{KV}}$.

\subsection{Prefill planning, migration, and admission}
For $n$ prompt tokens, the prefill target is $T_{\mathrm P}=B_{\mathrm P}nC_{\mathrm{full}}$, subject to sink/recent-window protection. The floor-constrained allocation uses the frozen priorities described in Section~\ref{sec:prefill-companion}; ties follow token position and then layer-group order. Floor excess is recorded rather than overriding protection.

The selected actions determine persistent counts
\[
H_c=\sum_{t=1}^{n}\mathbb I[a^{\mathrm P}_{t,c}=H],\qquad
L_c=\sum_{t=1}^{n}\mathbb I[a^{\mathrm P}_{t,c}=L],
\]
and payload $H_cm(H,c)+L_cm(L,c)$, including mask-protected positions. Reservations add slot/page rounding and metadata. Prefill first uses all-BF16 scratch storage, then retains $H$, quantizes $L$, and releases unprotected $R$.

Admission checks each pool against persistent demand, scratch storage, and overlapping migration buffers, deduplicating only allocator-supported aliases. Capacity shortage or allocation failure defers or retries a request.

\subsection{Packed attention and quantization}
The kernel reads disjoint high/low maps, separate K/V scales, current KV, and logical positions. GQA maps query heads to shared KV heads~\citep{ainslie-etal-2023-gqa}, supporting one-head, GQA-2, GQA-4, and split-KV routes. Each visible key appears once across persistent, recent, and current storage; absolute positions and causal masks survive storage gaps.

Each token--KV-head K/V row has an independent symmetric INT4 scale: maximum absolute value divided by 7, rounded to FP16 and bounded below by its smallest positive representable value. $P/s^P$ is rounded to the nearest integer and clipped to $[-7,7]$, using the same stored scale for dequantization. Zero rows use zero codes and unit scale. Two signed nibbles occupy one byte; odd dimensions add masked zero padding. Scales have shape $(\mathrm{slot},\mathrm{KV\ head},1)$.

Dequantization occurs inside attention. Query arithmetic, scores, softmax statistics, and value accumulation use FP32; output is BF16. Nonempty disjoint partitions with score maximum $v_j$, exponential sum $l_j$, and unnormalized value sum $u_j$ share normalization:
\begin{equation}
\begin{aligned}
v=\max_j v_j, \quad
o=\frac{\sum_j e^{v_j-v}u_j}{\sum_j e^{v_j-v}l_j}.
\end{aligned}
\label{eq:app-softmax-merge}
\end{equation}
Empty partitions contribute zero; independently normalized outputs are not averaged.

\subsection{Decode budget lifecycle and reserve}
\label{app:pask-budget-details}
The per-token allowance is $b=B_{\mathrm D}C_{\mathrm{full}}$ and retains fractional bits. The recurrence in Algorithm~\ref{alg:pask-online} equivalently gives $R_t=tb-\sum_{i<t}M(\mathbf a_i)$, with $M(\mathbf a)=\sum_c m(a_c,c)$. This balance is a soft scheduled-payload target, not a physical-memory cap. Floor-induced debt carries forward. The quantity $[M(\mathbf a_t)-R_t]_+$ measures the excess of the current action cost over the pre-action balance, which may already include debt from previous tokens.

For reference floor cost $f_{r,v}=\sum_c m(a_{\min}(q^{\mathrm{ref}}_{r,v,c}),c)$, the offline reserve statistic is
\begin{equation}
\begin{aligned}
S^{(K)}_{r,u}(B_{\mathrm D})
&=\max_{0\leq h\leq\min(K,T_r^{\mathrm{ref}}-u)}\\
&\quad\left[\sum_{j=1}^{h}f_{r,u+j}-hb\right]_+.
\end{aligned}
\label{eq:reserve-prefix-deficit}
\end{equation}
The prefix maximum captures temporary deficits that subsequent allowance could offset. Equation~\ref{eq:reserve} uses $K=8$ and $\beta=0.8$; the empirical quantile is the value at rank $\lceil\beta N\rceil$ among $N$ sorted samples, with a pooled fallback for unseen states. Future reference tokens enter only this offline construction. Calibration-distribution coverage does not imply coverage on unseen trajectories.

\subsection{Termination, rollback, and control complexity}
Stop tokens incur no new charge. The controller applies through the generation limit; termination reclaims request storage, and the task oracle scores incomplete outputs. Speculative proposals leave the committed ledger unchanged. Rollback restores parser state, budget, maps, and pending actions, undoing both allowances and charges. Destructive KV updates are deferred or snapshotted until commitment; restoring accounting alone cannot recover discarded KV.

After tag resolution, lookup and parent fallback use precomputed entries. At most two adjacent changes per group in either direction yield $O(|\mathcal C|\log|\mathcal C|)$ control work and $O(|\mathcal C|)$ state per token, independent of history length for fixed groups. This excludes frontier adaptation, quantization, attention, storage, and rollback history. Physical-capacity shortages defer decoding rather than violate floors.

\section{Experimental Details}
\label{app:experimental-details}

\subsection{Datasets}
BFCL uses a seeded $1{:}1{:}3$ calibration/development/test split within each non-Live and Live category, assigning complete normalized tool-call schemas to a single partition. For BIRD, we use the official 2024-06-27 Dev package, partitioned by database: \texttt{debit\_card\_specializing} and \texttt{european\_football\_2} for calibration (193 questions), \texttt{california\_schools} and \texttt{financial} for development (195), and the remaining seven databases for held-out evaluation (1,146). Table~\ref{tab:app-datasets} reports the category-level composition.

\begin{table*}[t]
\centering
\footnotesize
\setlength{\tabcolsep}{5pt}
\renewcommand{\arraystretch}{1.1}
\begin{tabular}{@{}llccp{0.5\textwidth}@{}}
\toprule
\textbf{Benchmark} & \textbf{Subset} &
\textbf{Cases} & \textbf{Test} &
\textbf{Description} \\
\midrule
\multirow{5}{*}{\shortstack[l]{BFCL non-live\\(cases 1,000)}}
 & Simple & 400 & 240 &
 A single tool call matching one target function schema. \\
 & Multiple & 200 & 120 &
 One tool call selected among several candidate functions. \\
 & Parallel & 200 & 120 &
 Several independent tool calls emitted in a single response. \\
 & Parallel-multiple & 200 & 120 &
 Several calls drawn from several candidate functions. \\
\cmidrule(lr){2-5}
 & All subsets & 1,000 & 600 &
 200 calibration and 200 development cases. \\
\midrule
\multirow{5}{*}{\shortstack[l]{BFCL Live\\(cases 1,351)}}
 & Simple & 258 & 154 &
 Real-user queries with a single tool call. \\
 & Multiple & 1,053 & 633 &
 Real-user queries requiring function selection among candidates. \\
 & Parallel & 16 & 10 &
 Real-user queries with several independent calls. \\
 & Parallel-multiple & 24 & 14 &
 Real-user queries with several calls over candidate functions. \\
\cmidrule(lr){2-5}
 & All subsets & 1,351 & 811 &
 270 calibration and 270 development cases. \\
\midrule
\multirow{4}{*}{\shortstack[l]{BIRD Dev\\(cases 1,534)}}
 & Simple & 925 & 701 &
 Straightforward SQL over the given database schema. \\
 & Moderate & 464 & 330 &
 Queries involving joins, aggregations, or nested conditions. \\
 & Challenging & 145 & 115 &
 Queries requiring complex reasoning over schemas and values. \\
\cmidrule(lr){2-5}
 & All databases & 1,534 & 1,146 &
 Seven held-out SQLite databases: toxicology (145),
 card\_games (191), codebase\_community (186),
 superhero (129), formula\_1 (174),
 thrombosis\_prediction (163), and student\_club (158). \\
\bottomrule
\end{tabular}
\vspace{-0.08in}
\caption{Benchmark composition. \textbf{Cases} denotes the full benchmark and \textbf{Test} the held-out evaluation subset.}
\label{tab:app-datasets}
\vspace{-0.08in}
\end{table*}

\subsection{Task-Accuracy Evaluation}
BFCL predictions are scored by the official AST checker, with Overall accuracy defined as the instance-weighted fraction of correct cases. BIRD execution accuracy (EX) uses the official \texttt{mini\_dev} evaluator. All methods share the same prompts, SGLang/XGrammar constrained-decoding backend, greedy decoding (temperature $0$), and task scorer. Results are deterministic single-run point estimates.

PASK uses these task oracles for the calibration interventions defined in Section~\ref{sec:offline-calibration}, whereas baseline task feedback is used for offline hyperparameter selection. No method receives task labels or task-error feedback during inference. Baseline implementations and their budget conventions are detailed in Appendix~\ref{app:baseline-tuning}.\footnote{Official implementations: H2O (\href{https://github.com/FMInference/H2O}{FMInference/H2O}), SnapKV (\href{https://github.com/FasterDecoding/SnapKV}{FasterDecoding/SnapKV}), RocketKV (\href{https://github.com/NVlabs/RocketKV}{NVlabs/RocketKV}), LazyEviction (\href{https://github.com/Halo-949/LazyEviction}{Halo-949/LazyEviction}), and CriticalKV (via the authors' \href{https://github.com/FFY0/AdaKV}{FFY0/AdaKV}). No standalone official public repositories were available at the time of writing for MiKV, TriAxialKV, or Transactional Attention.}

\subsection{Baseline tuning and implementation}
\label{app:baseline-tuning}

\textbf{Tuning protocol.}
Each baseline is tuned separately for every model--dataset pair: BFCL non-Live and Live with Qwen3-4B, Qwen3-14B, and LLaMA-3.1-8B, and BIRD with Qwen3-4B. Candidate configurations are selected on the development split using the official task metric and frozen before test evaluation.

Budget-flexible methods are adjusted to the target KV budget. Fixed-profile methods that cannot exactly match the target retain their native allocation and precision profiles, with measured KV payload ratios reported separately. Table~\ref{tab:baseline-search} lists representative high-impact search dimensions; secondary implementation-specific parameters are omitted.

\begin{table*}[t]
\centering
\scriptsize
\setlength{\tabcolsep}{4pt}
\renewcommand{\arraystretch}{1.10}
\begin{tabular}{@{}p{0.15\textwidth}
                    p{0.2\textwidth}
                    p{0.6\textwidth}@{}}
\toprule
\textbf{Baseline} &
\textbf{Representative tuned knobs} &
\textbf{Candidate values} \\
\midrule

H2O &
Sink size; recent window &
Sink $\in \{2,4,8\}$;
recent window $\in \{16,32,64,128\}$. \\

SnapKV &
Observation window; pooling; pooling kernel &
Observation window $\in \{16,32,64,128\}$;
pooling $\in \{\mathrm{avg},\mathrm{max}\}$;
kernel $\in \{3,5,7\}$. \\

MiKV &
High-precision ratio; low-bit width; group size &
High-precision ratio $\in \{0.10,0.20,0.30,0.40\}$;
low-bit width $\in \{2,4\}$;
group size $\in \{32,64,128\}$. \\

RocketKV &
Observation window; pooling kernel; prefill ratio &
Observation window $\in \{16,32,64,128\}$;
kernel $\in \{31,63,127\}$;
prefill ratio $\in \{0.20,0.25,0.30,0.35\}$. \\

CriticalKV &
First-stage ratio; stabilizer $\epsilon$; scorer window &
First-stage ratio $\in \{0.25,0.40,0.50,0.70\}$;
$\epsilon \in \{10^{-5},10^{-4},10^{-3}\}$;
scorer window $\in \{16,32,64\}$. \\

LazyEviction &
Observation/recent window; attention threshold $\alpha$ &
Observation/recent window $\in \{8,16,32,64\}$;
$\alpha \in \{10^{-5},10^{-4},3{\times}10^{-4},10^{-3}\}$. \\

TriAxialKV &
Protected prefix; precision profile; recent window &
Protected prefix $\in \{4,8,16,32\}$;
(low, high) bits $\in \{(2,4),(3,4),(4,4),(2,8)\}$;
recent window $\in \{32,128,256\}$. \\

Transactional Attention &
Anchor mode; sponsor span; protected fraction &
Anchor mode $\in \{\mathrm{regex},\mathrm{semantic}\}$;
sponsor span $\in \{4,6,8,12\}$;
protected fraction $\in \{0.15,0.25,0.35\}$. \\

\bottomrule
\end{tabular}
\vspace{-0.08in}
\caption{Representative baseline hyperparameter search spaces. }
\label{tab:baseline-search}
\vspace{-0.08in}
\end{table*}

\textbf{H2O.}
H2O accumulates attention received by past tokens across decoding steps and retains the highest-scoring history together with protected sink and recent tokens.

\textbf{SnapKV.}
SnapKV uses attention from the final prompt queries to score prefix tokens and pools neighboring scores before selecting the retained prefix.

\textbf{MiKV.}
MiKV assigns importance-ranked KV entries to a high-precision tier and stores the remaining entries at lower precision rather than dropping them. We use H2O-style cumulative-attention importance. No author implementation was available when this reproduction was built. The high-precision ratio is tuned separately at each target budget.

\textbf{RocketKV.}
RocketKV combines pooled attention-based prompt selection with coarse-to-fine sparse attention during decoding. 

We use a batched SGLang implementation of its two-stage algorithm. Prefill selection scores prefix tokens with the observation-window scorer (odd max-pool kernel 63) and retains $\lceil r_{\mathrm{prompt}}\cdot L\rceil$ tokens,
where $r_{\mathrm{prompt}}$ is the configured prompt-retention ratio under the same shared-token-mask convention as SnapKV. During decoding, newly generated KV remains in BF16, while sparse attention reads $\lceil 0.5\cdot\text{prompt budget}\rceil$ selected tokens per request and KV head using the chunked max-key sketch with head- and sequence-dimension reductions. The official fused kernels are integrated with TensorRT-LLM and are not directly composable with our constrained-generation stack; therefore, the same-stack comparison uses this batched implementation.

\textbf{CriticalKV.}
CriticalKV first constructs a candidate set using a SnapKV-style attention score and then reranks candidates according to the output perturbation induced by removing individual value vectors.

\textbf{LazyEviction.}
LazyEviction tracks recurring attention over an observation window and delays eviction so that tokens that regain importance can remain cached.

\textbf{TriAxialKV.}
Our text-only reproduction follows the per-tag attention-output distortion calibration and budget-constrained INT2/INT4 allocation procedure of TriAxialKV~\citep{shen2026triaxialkvextremelowprecisionkvcache}. It assigns precision according to semantic tags, protecting the prefix and recent tail while using lower precision for older prompt tokens. We freeze the calibrated allocation before test evaluation and report its measured KV cost separately, since the evaluated configuration does not match the target budget. Under this configuration, TriAxialKV uses an element-weighted average precision of 3.84 bits for stored K/V values and has a measured relative total KV cost of $B=0.24$.

\textbf{Transactional Attention.}
TA-Fast-Semantic detects semantic anchors and assigns decaying sponsorship vouchers to nearby tokens, which contribute to the token-retention utility score. Additional sponsorship and utility parameters are tuned under the same held-out selection protocol but are omitted from Table~\ref{tab:baseline-search} for compactness.

\subsection{System experiment}

The serving stack uses SGLang 0.4.6.post5, PyTorch 2.6.0+cu124 and XGrammar 0.1.21. Radix caching, overlap scheduling, and chunked prefill are disabled in the primary serving matrix.

Each GPU configuration is measured over three successful repetitions, with medians reported. Concurrency $c$ denotes requests submitted concurrently in each client-side wave, including those queued for server admission; checkpoint metrics are collected at the specified generated-token position using matched request-level workloads.

\paragraph{Resource licenses.}
Our code release contains original PASK code and modified SGLang source files. The modified SGLang files are redistributed under Apache License 2.0, with original license and attribution notices retained and modifications explicitly marked. Qwen3-4B-Instruct-2507 and Qwen3-14B are distributed under Apache License 2.0. LLaMA-3.1-8B-Instruct is distributed under the Llama 3.1 Community License Agreement and is subject to its Acceptable Use Policy. BFCL is distributed under Apache License 2.0, while BIRD's published data license is Creative Commons Attribution-ShareAlike 4.0 International (CC BY-SA 4.0). We use these models and datasets for offline research evaluation and do not redistribute benchmark datasets or pretrained model weights.

\begin{table*}[t]
\centering
\footnotesize
\setlength{\tabcolsep}{3pt}
\renewcommand{\arraystretch}{1}
\resizebox{\textwidth}{!}{%
\begin{tabular}{@{}lc|ccccc|ccccc@{}}
\toprule
\multirow{2}{*}{\textbf{Method}} & \multirow{2}{*}{\textbf{B}} & \multicolumn{5}{c|}{\textbf{BFCL non-live}} & \multicolumn{5}{c}{\textbf{BFCL live}} \\
\cmidrule(lr){3-7}\cmidrule(lr){8-12}
& & \textbf{Simple} & \textbf{Multiple} & \textbf{Parallel} & \textbf{Par.-multiple} & \textbf{Overall} & \textbf{Simple} & \textbf{Multiple} & \textbf{Parallel} & \textbf{Par.-multiple} & \textbf{Overall} \\
\midrule
Full KV & 1.00 & 95.00 & 95.00 & 90.00 & 84.17 & 91.83 & 78.57 & 74.57 & 60.00 & 78.57 & 75.22 \\
\midrule
H2O & 0.33 & 75.83 & 81.67 & 31.67 & 19.17 & 56.83 & 45.45 & 48.18 & 0.00 & 21.43 & 46.61 \\
SnapKV & 0.33 & 85.00 & 87.50 & 42.50 & 35.00 & 67.00 & 55.19 & 59.08 & 0.00 & 14.29 & 56.84 \\
MiKV & 0.33 & 87.50 & 90.83 & 76.67 & 76.67 & 83.83 & 64.94 & 64.45 & 30.00 & 50.00 & 63.87 \\
RocketKV & 0.33 & 75.42 & 65.83 & 33.33 & 25.00 & 55.00 & 48.05 & 47.24 & 0.00 & 28.57 & 46.49 \\
TransAttn & 0.33 & 88.75 & 64.17 & 54.17 & 43.33 & 67.83 & 45.45 & 27.49 & 0.00 & 0.00 & 30.09 \\
CriticalKV & 0.33& 87.92 & 91.67 & 59.17 & 40.00 & 73.33 & 59.74 & 66.19 & 30.00 & 28.57 & 63.87 \\
LazyEviction  & 0.33 & 46.67 & 81.67 & 20.83 & 14.17 & 42.00 & 33.12 & 60.03 & 10.00 & 28.57 & 53.76 \\
TriAxialKV & fixed & 90.42 & 85.00 & 70.83 & 56.67 & 78.67 & 62.99 & 59.08 & 40.00 & 0.00 & 58.57 \\
\cellcolor{oursrow}\textbf{PASK (Ours)} & \cellcolor{oursrow}\textbf{0.33} & \cellcolor{oursrow}\textbf{92.08} & \cellcolor{oursrow}\textbf{93.33} & \cellcolor{oursrow}\textbf{80.83} & \cellcolor{oursrow}\textbf{81.67} & \cellcolor{oursrow}\textbf{88.00} & \cellcolor{oursrow}\textbf{74.03} & \cellcolor{oursrow}\textbf{72.20} & \cellcolor{oursrow}\textbf{60.00} & \cellcolor{oursrow}\textbf{50.00} & \cellcolor{oursrow}\textbf{72.01} \\
\midrule
H2O & 0.40 & 77.92 & 84.17 & 42.50 & 34.17 & 63.33 & 50.00 & 51.97 & 0.00 & 21.43 & 50.43 \\
SnapKV & 0.40 & 89.17 & 94.17 & 45.00 & 39.17 & 71.33 & 59.74 & 69.98 & 20.00 & 28.57 & 66.71 \\
MiKV & 0.40 & 88.33 & 90.83 & 79.17 & 75.00 & 84.33 & 66.23 & 64.61 & 30.00 & 42.86 & 64.12\\
RocketKV & 0.40 & 86.67 & 95.83 & 65.00 & 41.67 & 75.17 & 59.74 & 63.03 & 20.00 & 35.71 & 61.41 \\
TransAttn & 0.40 & 90.83 & 73.33 & 61.67 & 47.50 & 72.83 & 52.60 & 44.71 & 20.00 & 0.00 & 45.13 \\
CriticalKV & 0.40 & 91.25 & 92.50 & 61.67 & 64.17 & 80.17& 60.39 & 72.04 & 40.00 & 35.71 & 68.80  \\
LazyEviction  & 0.40 & 58.75 & 76.67 & 34.17 & 38.33 & 53.33 & 47.40 & 61.77 & 10.00 & 50.00 & 58.20  \\
\cellcolor{oursrow}\textbf{PASK (Ours)} & \cellcolor{oursrow}\textbf{0.40} & \cellcolor{oursrow}\textbf{95.83} & \cellcolor{oursrow}\textbf{97.50} & \cellcolor{oursrow}\textbf{86.67} & \cellcolor{oursrow}\textbf{80.00} & \cellcolor{oursrow}\textbf{91.17} & \cellcolor{oursrow}\textbf{77.27} & \cellcolor{oursrow}\textbf{74.88} & \cellcolor{oursrow}\textbf{50.00} & \cellcolor{oursrow}\textbf{50.00} & \cellcolor{oursrow}\textbf{74.60} \\
\midrule
H2O & 0.50 & 81.25 & 85.83 & 52.50 & 40.00 & 68.17 & 52.60 & 60.66 & 10.00 & 28.57 & 57.95 \\
SnapKV & 0.50 & 92.08 & 95.83 & 66.67 & 61.67 & 81.67 & 62.34 & 74.57 & 50.00 & 28.57 & 71.15 \\
MiKV & 0.50 & 89.17 & 90.83 & 80.00 & 73.33 & 84.50 & 69.48 & 65.09 & 30.00 & 42.86 & 65.10\\
RocketKV & 0.50 & 90.00 & 94.17 & 75.83 & 60.83 & 82.17 & 58.44 & 69.67 & 30.00 & 42.86 & 66.58 \\
TransAttn & 0.50 & 94.58 & 78.33 & 85.83 & 62.50 & 83.17 & 68.18 & 46.76 & 60.00 & 21.43 & 50.55 \\
CriticalKV & 0.50& 92.92 & 95.00 & 75.00 & 70.00 & 85.17 & 64.29 & 74.57 & 50.00 & 42.86 & 71.76 \\
LazyEviction  & 0.50 & 75.42 & 90.83 & 35.00 & 35.83 & 62.50 & 55.19 & 68.56 & 10.00 & 57.14 & 65.10 \\
\cellcolor{oursrow}\textbf{PASK (Ours)} & \cellcolor{oursrow}\textbf{0.50} & \cellcolor{oursrow}\textbf{95.00} & \cellcolor{oursrow}\textbf{97.50} & \cellcolor{oursrow}\textbf{87.50} & \cellcolor{oursrow}\textbf{81.67} & \cellcolor{oursrow}\textbf{91.33} & \cellcolor{oursrow}\textbf{79.22} & \cellcolor{oursrow}\textbf{74.72} & \cellcolor{oursrow}\textbf{50.00} & \cellcolor{oursrow}\textbf{57.14} & \cellcolor{oursrow}\textbf{74.97} \\
\midrule
H2O & 0.60 & 85.42 & 90.00 & 65.00 & 58.33 & 76.83 & 57.14 & 64.14 & 20.00 & 35.71 & 61.78 \\
SnapKV & 0.60 & 93.33 & 96.67 & 83.33 & 78.33 & 89.00 & 69.48 & 75.36 & 50.00 & 42.86 & 73.37 \\
MiKV & 0.60 & 92.08 & 91.67 & 83.33 & 75.00 & 86.83 & 71.43 & 66.51 & 50.00 & 71.43 & 67.32\\
RocketKV & 0.60 & 90.42 & 94.17 & 80.83 & 75.83 & 86.33 & 62.99 & 72.83 & 30.00 & 35.71 & 69.79 \\
TransAttn & 0.60 & 95.42 & 95.00 & 89.17 & 74.17 & 89.83 & 74.68 & 65.72 & 60.00 & 57.14 & 67.20 \\
CriticalKV & 0.60 & 94.58 & 95.00 & 80.00 & 81.67 & 89.17 & 69.48 & 74.88 & 50.00 & 57.14 & 73.24   \\
LazyEviction  & 0.60 & 87.08 & 83.33 & 35.83 & 45.00 & 67.67 & 52.60 & 76.15 & 0.00 & 57.14 & 70.41 \\
\cellcolor{oursrow}\textbf{PASK (Ours)} & \cellcolor{oursrow}\textbf{0.60} & \cellcolor{oursrow}\textbf{95.42} & \cellcolor{oursrow}\textbf{97.50} & \cellcolor{oursrow}\textbf{85.83} & \cellcolor{oursrow}\textbf{84.17} & \cellcolor{oursrow}\textbf{91.67} & \cellcolor{oursrow}\textbf{77.92} & \cellcolor{oursrow}\textbf{74.88} & \cellcolor{oursrow}\textbf{60.00} & \cellcolor{oursrow}\textbf{64.29} & \cellcolor{oursrow}\textbf{75.09} \\
\bottomrule
\end{tabular}%
}
\caption{Accuracy (\%) at multiple KV budgets on Qwen3-4B. }
\vspace{-0.1in}
\label{tab:qwen4b-results}
\end{table*}

\subsection{Structural ablation protocols.}
The canonical policy key combines token class, structural role, parser state, lookahead, and layer group. The no-class and no-role variants remove their respective components, while no-state removes both parser state and lookahead. The no-layer variant ties actions across the three layer groups without removing any physical layers.

The no-parser-transition variant replaces structural tags with request-local position bins, retaining layer groups and the $R/L/H$ action set. We construct ten contiguous bins from Full-KV calibration completion lengths, requiring each bin to be reached by at least 75\% of calibration requests in every BFCL category; the final bin absorbs the tail. The parser enforces token legality but does not inform KV-action lookup.

For each variant, we reconstruct its policy buckets and independently recalibrate Runtime-V task-error sensitivity and Runtime-D attention-output distortion using the same calibration split and single-bucket intervention protocol. Each variant uses its measurements and policy.

\section{Additional Evidence}
\label{app:additional-evidence}
\subsection{Multi-budget BFCL results}
\label{app:multibudget-accuracy}
Tables~\ref{tab:qwen4b-results}--\ref{tab:llama31-results} provide category-level BFCL results underlying Fig.~\ref{fig:bfcl-budget-grid}.

\begin{table*}[t]
\centering
\footnotesize
\setlength{\tabcolsep}{3pt}
\renewcommand{\arraystretch}{1}
\resizebox{\textwidth}{!}{%
\begin{tabular}{@{}lc|ccccc|ccccc@{}}
\toprule
\multirow{2}{*}{\textbf{Method}} & \multirow{2}{*}{\textbf{B}} & \multicolumn{5}{c|}{\textbf{BFCL non-live}} & \multicolumn{5}{c}{\textbf{BFCL live}} \\
\cmidrule(lr){3-7}\cmidrule(lr){8-12}
& & \textbf{Simple} & \textbf{Multiple} & \textbf{Parallel} & \textbf{Par.-multiple} & \textbf{Overall} & \textbf{Simple} & \textbf{Multiple} & \textbf{Parallel} & \textbf{Par.-multiple} & \textbf{Overall} \\
\midrule
Full KV & 1.00 & 96.25 & 98.33 & 92.50 & 81.67 & 93.00 & 79.87 & 75.20 & 70.00 & 64.29 & 75.83 \\
\midrule
H2O & 0.33 & 72.08 & 85.00 & 22.50 & 16.67 & 53.67 & 54.55 & 51.82 & 0.00 & 0.00 & 50.80 \\
SnapKV & 0.33 & 82.50 & 91.67 & 60.00 & 55.83 & 74.50 & 61.69 & 66.98 & 0.00 & 14.29 & 64.24 \\
MiKV & 0.33 & 83.75 & 88.33 & 79.17 & 65.83 & 80.17 & 66.23 & 63.03 & 40.00 & 42.86 & 63.01 \\
RocketKV & 0.33 & 78.33 & 90.00 & 30.00 & 41.67 & 63.67 & 55.19 & 55.13 & 30.00 & 42.86 & 54.62 \\
TransAttn & 0.33 & 88.33 & 83.33 & 70.00 & 49.17 & 75.83 & 49.35 & 31.12 & 0.00 & 0.00 & 33.66 \\
CriticalKV & 0.33 & 80.42 & 87.50 & 54.17 & 50.83 & 70.67 & 61.69 & 65.72 & 40.00 & 57.14 & 64.49 \\
LazyEviction  & 0.33 & 49.17 & 91.67 & 28.33 & 25.00 & 48.67 & 33.77 & 64.14 & 20.00 & 57.14 & 57.71 \\
TriAxialKV & fixed & 87.92 & 79.17 & 77.50 & 64.17 & 79.33 & 68.18 & 61.14 & 30.00 & 0.00 & 61.04 \\
\cellcolor{oursrow}\textbf{PASK (Ours)} & \cellcolor{oursrow}\textbf{0.33} & \cellcolor{oursrow}\textbf{92.08} & \cellcolor{oursrow}\textbf{94.17} & \cellcolor{oursrow}\textbf{86.67} & \cellcolor{oursrow}\textbf{79.17} & \cellcolor{oursrow}\textbf{88.83} & \cellcolor{oursrow}\textbf{73.38} & \cellcolor{oursrow}\textbf{73.62} & \cellcolor{oursrow}\textbf{30.00} & \cellcolor{oursrow}\textbf{57.14} & \cellcolor{oursrow}\textbf{72.75} \\
\midrule
H2O & 0.40 & 77.08 & 87.50 & 40.00 & 35.00 & 63.33 & 57.79 & 57.35 & 10.00 & 21.43 & 56.23 \\
SnapKV & 0.40 & 82.08 & 89.17 & 65.83 & 60.00 & 75.83 & 64.94 & 70.30 & 30.00 & 50.00 & 68.43 \\
MiKV & 0.40 & 88.75&92.50&85.00&70.83&85.17&65.58&63.67&40.00&42.86&63.38\\
RocketKV & 0.40 & 84.58 & 94.17 & 51.67 & 41.67 & 71.33 & 57.14 & 65.88 & 30.00 & 57.14 & 63.63 \\
TransAttn & 0.40 & 91.67 & 85.00 & 73.33 & 46.67 & 77.67 & 55.84 & 45.34 & 40.00 & 21.43 & 46.86 \\
CriticalKV & 0.40 & 81.67 & 90.00 & 60.00 & 70.83 & 76.83 & 62.34 & 69.67 & 40.00 & 64.29 & 67.82   \\
LazyEviction  & 0.40 & 62.50 & 86.67 & 30.83 & 36.67 & 55.83 & 50.00 & 66.19 & 0.00 & 50.00 & 62.02   \\
\cellcolor{oursrow}\textbf{PASK (Ours)} & \cellcolor{oursrow}\textbf{0.40} & \cellcolor{oursrow}\textbf{93.33} & \cellcolor{oursrow}\textbf{94.17} & \cellcolor{oursrow}\textbf{87.50} & \cellcolor{oursrow}\textbf{78.33} & \cellcolor{oursrow}\textbf{89.33} & \cellcolor{oursrow}\textbf{75.32} & \cellcolor{oursrow}\textbf{74.09} & \cellcolor{oursrow}\textbf{40.00} & \cellcolor{oursrow}\textbf{57.14} & \cellcolor{oursrow}\textbf{73.61} \\
\midrule
H2O & 0.50 & 82.92 & 89.17 & 57.50 & 48.33 & 72.17 & 59.09 & 62.72 & 20.00 & 28.57 & 60.91 \\
SnapKV & 0.50 & 85.83 & 93.33 & 75.83 & 71.67 & 82.50 & 64.29 & 71.09 & 40.00 & 50.00 & 69.05 \\
MiKV & 0.50 &88.75&91.67&85.83&71.67&85.33&64.94&64.77&40.00&42.86&64.12\\
RocketKV & 0.50 & 86.25 & 94.17 & 73.33 & 59.17 & 79.83 & 60.39 & 67.77 & 40.00 & 57.14 & 65.84 \\
TransAttn & 0.50 & 95.42 & 88.33 & 89.17 & 58.33 & 85.33 & 64.29 & 48.50 & 30.00 & 28.57 & 50.92 \\
CriticalKV & 0.50 & 85.00 & 95.00 & 83.33 & 74.17 & 84.50 & 66.23 & 71.09 & 40.00 & 71.43 & 69.79 \\
LazyEviction  & 0.50 & 70.00 & 94.17 & 49.17 & 46.67 & 66.00 & 52.60 & 69.35 & 20.00 & 57.14 & 65.35 \\
\cellcolor{oursrow}\textbf{PASK (Ours)} & \cellcolor{oursrow}\textbf{0.50} & \cellcolor{oursrow}\textbf{93.33} & \cellcolor{oursrow}\textbf{94.17} & \cellcolor{oursrow}\textbf{88.33} & \cellcolor{oursrow}\textbf{80.83} & \cellcolor{oursrow}\textbf{90.00} & \cellcolor{oursrow}\textbf{76.62} & \cellcolor{oursrow}\textbf{74.41} & \cellcolor{oursrow}\textbf{50.00} & \cellcolor{oursrow}\textbf{57.14} & \cellcolor{oursrow}\textbf{74.23} \\
\midrule
H2O & 0.60 & 85.83 & 90.83 & 74.17 & 62.50 & 79.83 & 61.69 & 66.35 & 30.00 & 50.00 & 64.73 \\
SnapKV & 0.60 & 86.67 & 95.00 & 85.00 & 79.17 & 86.50 & 62.99 & 71.56 & 40.00 & 57.14 & 69.30 \\
MiKV & 0.60 & 89.58&92.50&88.33&76.67&87.33&65.58&66.51&70.00&50.00&66.09 \\
RocketKV & 0.60 & 89.17 & 93.33 & 82.50 & 70.00 & 84.83 & 63.64 & 69.67 & 40.00 & 64.29 & 68.06 \\
TransAttn & 0.60 & 95.83 & 98.33 & 91.67 & 75.83 & 91.50 & 72.73 & 51.66 & 30.00 & 28.57 & 54.99 \\
CriticalKV & 0.60 & 90.00 & 95.83 & 85.83 & 84.17 & 89.17 & 72.73 & 72.20 & 60.00 & 50.00 & 71.76  \\
LazyEviction  & 0.60 & 90.83 & 89.17 & 51.67 & 52.50 & 75.00 & 63.64 & 74.57 & 10.00 & 35.71 & 71.02 \\
\cellcolor{oursrow}\textbf{PASK (Ours)} & \cellcolor{oursrow}\textbf{0.60} & \cellcolor{oursrow}\textbf{95.00} & \cellcolor{oursrow}\textbf{97.50} & \cellcolor{oursrow}\textbf{94.17} & \cellcolor{oursrow}\textbf{83.33} & \cellcolor{oursrow}\textbf{93.00} & \cellcolor{oursrow}\textbf{78.57} & \cellcolor{oursrow}\textbf{74.88} & \cellcolor{oursrow}\textbf{60.00} & \cellcolor{oursrow}\textbf{57.14} & \cellcolor{oursrow}\textbf{75.09} \\
\bottomrule
\end{tabular}%
}
\caption{Accuracy (\%) at multiple KV budgets on Qwen3-14B.}
\vspace{-0.1in}
\label{tab:qwen14b-results}
\end{table*}

\begin{table*}[t]
\centering
\footnotesize
\setlength{\tabcolsep}{3pt}
\renewcommand{\arraystretch}{1}
\resizebox{\textwidth}{!}{%
\begin{tabular}{@{}lc|ccccc|ccccc@{}}
\toprule
\multirow{2}{*}{\textbf{Method}} & \multirow{2}{*}{\textbf{B}} & \multicolumn{5}{c|}{\textbf{BFCL non-live}} & \multicolumn{5}{c}{\textbf{BFCL live}} \\
\cmidrule(lr){3-7}\cmidrule(lr){8-12}
& & \textbf{Simple} & \textbf{Multiple} & \textbf{Parallel} & \textbf{Par.-multiple} & \textbf{Overall} & \textbf{Simple} & \textbf{Multiple} & \textbf{Parallel} & \textbf{Par.-multiple} & \textbf{Overall} \\
\midrule
Full KV & 1.00 & 94.58 & 95.00 & 91.67 & 80.83 & 91.33 & 75.32 & 60.98 & 40.00 & 57.14 & 63.38 \\
\midrule
H2O & 0.33 & 75.42 & 90.00 & 15.83 & 12.50 & 53.83 & 46.75 & 36.65 & 0.00 & 14.29 & 37.73 \\
SnapKV & 0.33 & 87.92 & 94.17 & 43.33 & 35.00 & 69.67 & 51.30 & 53.55 & 10.00 & 35.71 & 52.28 \\
MiKV & 0.33 & 91.67 & 95.00 & 66.67 & 51.67 & 79.33 & 64.94 & 48.50 & 20.00 & 28.57 & 50.92 \\
RocketKV & 0.33 & 82.92 & 90.83 & 25.00 & 21.67 & 60.67 & 51.30 & 47.39 & 0.00 & 21.43 & 47.10 \\
TransAttn & 0.33 & 90.00 & 67.50 & 71.67 & 38.33 & 71.50 & 60.39 & 32.07 & 30.00 & 0.00 & 36.87 \\
CriticalKV & 0.33 & 86.67 & 93.33 & 35.83 & 33.33 & 67.17 & 62.34 & 53.08 & 10.00 & 35.71 & 54.01 \\
LazyEviction  & 0.33 & 56.67 & 88.33 & 25.83 & 16.67 & 48.83 & 38.96 & 45.66 & 10.00 & 28.57 & 43.65 \\
TriAxialKV & fixed & 92.08 & 88.33 & 85.83 & 55.83 & 82.83 & 66.23 & 38.23 & 50.00 & 28.57 & 43.53 \\
\cellcolor{oursrow}\textbf{PASK (Ours)} & \cellcolor{oursrow}\textbf{0.33} & \cellcolor{oursrow}\textbf{93.75} & \cellcolor{oursrow}\textbf{96.67} & \cellcolor{oursrow}\textbf{90.83} & \cellcolor{oursrow}\textbf{80.83} & \cellcolor{oursrow}\textbf{91.17} & \cellcolor{oursrow}\textbf{73.38} & \cellcolor{oursrow}\textbf{58.45} & \cellcolor{oursrow}\textbf{40.00} & \cellcolor{oursrow}\textbf{50.00} & \cellcolor{oursrow}\textbf{60.91} \\
\midrule
H2O & 0.40 & 78.75 & 90.83 & 28.33 & 25.83 & 60.50 & 49.35 & 41.86 & 0.00 & 14.29 & 42.29 \\
SnapKV & 0.40 & 90.83 & 94.17 & 50.00 & 38.33 & 72.83 & 62.99 & 57.35 & 20.00 & 35.71 & 57.58 \\
MiKV & 0.40 &91.67&94.17&69.17&53.33&80.00&64.94&51.50&20.00&35.71&53.39\\
RocketKV & 0.40 & 82.92 & 90.83 & 25.00 & 21.67 & 60.67 & 63.64 & 51.50 & 10.00 & 28.57 & 52.90 \\
TransAttn & 0.40 & 93.33 & 69.17 & 80.00 & 36.67 & 74.50 & 59.09 & 40.44 & 30.00 & 0.00 & 43.16 \\
CriticalKV & 0.40 & 87.92 & 94.17 & 65.00 & 39.17 & 74.83 & 64.94 & 56.08 & 20.00 & 35.71 & 56.97  \\
LazyEviction  & 0.40 & 66.25 & 87.50 & 33.33 & 36.67 & 58.00 & 37.01 & 53.08 & 20.00 & 50.00 & 49.57 \\
\cellcolor{oursrow}\textbf{PASK (Ours)} & \cellcolor{oursrow}\textbf{0.40} & \cellcolor{oursrow}\textbf{93.75} & \cellcolor{oursrow}\textbf{95.00} & \cellcolor{oursrow}\textbf{90.83} & \cellcolor{oursrow}\textbf{80.83} & \cellcolor{oursrow}\textbf{90.83} & \cellcolor{oursrow}\textbf{73.38} & \cellcolor{oursrow}\textbf{59.72} & \cellcolor{oursrow}\textbf{40.00} & \cellcolor{oursrow}\textbf{57.14} & \cellcolor{oursrow}\textbf{62.02} \\
\midrule
H2O & 0.50 & 84.58 & 92.50 & 43.33 & 33.33 & 67.67 & 53.90 & 46.29 & 0.00 & 14.29 & 46.61 \\
SnapKV & 0.50 & 92.50 & 95.83 & 70.00 & 53.33 & 80.83 & 66.23 & 59.24 & 30.00 & 28.57 & 59.68 \\
MiKV & 0.50 & 91.67&93.33&70.00&63.33&82.00&64.94&53.40&10.00&35.71&54.75\\
RocketKV & 0.50 & 89.17 & 95.00 & 75.00 & 49.17 & 79.50 & 62.34 & 54.34 & 30.00 & 50.00 & 55.49 \\
TransAttn & 0.50 & 95.00 & 90.00 & 88.33 & 52.50 & 84.17 & 69.48 & 42.81 & 40.00 & 7.14 & 47.23 \\
CriticalKV & 0.50 & 92.50 & 98.33 & 67.50 & 52.50 & 80.67 & 67.53 & 59.08 & 30.00 & 28.57 & 59.80 \\
LazyEviction  & 0.50 & 80.83 & 94.17 & 48.33 & 41.67 & 69.17 & 54.55 & 53.71 & 0.00 & 42.86 & 53.02 \\
\cellcolor{oursrow}\textbf{PASK (Ours)} & \cellcolor{oursrow}\textbf{0.50} & \cellcolor{oursrow}\textbf{94.58} & \cellcolor{oursrow}\textbf{95.00} & \cellcolor{oursrow}\textbf{90.83} & \cellcolor{oursrow}\textbf{80.83} & \cellcolor{oursrow}\textbf{91.17} & \cellcolor{oursrow}\textbf{75.32} & \cellcolor{oursrow}\textbf{59.56} & \cellcolor{oursrow}\textbf{40.00} & \cellcolor{oursrow}\textbf{57.14} & \cellcolor{oursrow}\textbf{62.27} \\
\midrule
H2O & 0.60 & 89.58 & 93.33 & 53.33 & 48.33 & 74.83 & 58.44 & 50.39 & 10.00 & 21.43 & 50.92 \\
SnapKV & 0.60 & 94.17 & 95.83 & 70.00 & 59.17 & 82.67 & 68.83 & 62.09 & 30.00 & 35.71 & 62.52 \\
MiKV & 0.60 & 92.08 & 95.00 & 70.83 & 69.17 & 83.83 & 65.58 & 54.66 & 50.00 & 57.14 & 56.72 \\
RocketKV & 0.60 & 91.25 & 95.83 & 81.67 & 63.33 & 84.67 & 63.64 & 57.82 & 20.00 & 28.57 & 57.95 \\
TransAttn & 0.60 & 95.83 & 91.67 & 90.83 & 80.00 & 90.83 & 74.03 & 57.66 & 40.00 & 42.86 & 60.30 \\
CriticalKV & 0.60 & 95.00 & 95.83 & 70.00 & 65.00 & 84.17 & 74.68 & 60.19 & 30.00 & 57.14 & 62.52  \\
LazyEviction  & 0.60 & 94.17 & 95.00 & 50.00 & 55.83 & 77.83 & 41.56 & 63.35& 0.00 & 57.14 & 58.32\\
\cellcolor{oursrow}\textbf{PASK (Ours)} & \cellcolor{oursrow}\textbf{0.60} & \cellcolor{oursrow}\textbf{93.75} & \cellcolor{oursrow}\textbf{96.67} & \cellcolor{oursrow}\textbf{91.67} & \cellcolor{oursrow}\textbf{80.83} & \cellcolor{oursrow}\textbf{91.33} & \cellcolor{oursrow}\textbf{74.03} & \cellcolor{oursrow}\textbf{60.66} & \cellcolor{oursrow}\textbf{40.00} & \cellcolor{oursrow}\textbf{50.00} & \cellcolor{oursrow}\textbf{62.76} \\
\bottomrule
\end{tabular}%
}
\caption{Accuracy (\%) at multiple KV budgets on LLaMA-3.1-8B.}
\vspace{-0.1in}
\label{tab:llama31-results}
\end{table*}

\subsection{BIRD detailed  results}
\label{app:bird-difficulty-results}

Table~\ref{tab:app-bird-difficulty} decomposes BIRD execution accuracy by difficulty tier across the evaluated KV budgets.
\begin{table*}[t]
\centering
\footnotesize
\setlength{\tabcolsep}{6pt}
\renewcommand{\arraystretch}{1.05}
\begin{tabular}{@{}lccccc@{}}
\toprule
\textbf{Method} & \textbf{Budget} & \textbf{Simple EX} & \textbf{Moderate EX} & \textbf{Challenging EX} & \textbf{Overall EX} \\
\midrule
Full KV & 1.000 & 62.77 & 46.67 & 44.35 & 56.28 \\
\midrule
H2O & 0.275 & 32.81 & 10.61 & 9.57 & 24.08 \\
SnapKV & 0.275 & 44.79 & 25.15 & 19.13 & 36.56 \\
MiKV & 0.275 & 55.63 & 37.27 & 34.78 & 48.25 \\
RocketKV & 0.275 & 50.36 & 29.70 & 17.39 & 41.10 \\
CriticalKV & 0.275 & 45.22 & 28.18 & 17.39 & 37.52 \\
LazyEviction & 0.275 & 57.92 & 39.09 & 27.83 & 49.48 \\
TransAttn & 0.275 & 50.07 & 29.09 & 11.30 & 40.14 \\
\cellcolor{oursrow}\textbf{PASK (Ours)} & \cellcolor{oursrow}\textbf{0.275} & \cellcolor{oursrow}\textbf{60.77} & \cellcolor{oursrow}\textbf{41.52} & \cellcolor{oursrow}\textbf{35.65} & \cellcolor{oursrow}\textbf{52.71} \\
\midrule
H2O & 0.4 & 42.80 & 15.45 & 19.13 & 32.55 \\
SnapKV & 0.4 & 51.64 & 35.76 & 23.48 & 44.24 \\
MiKV & 0.4 & 58.35 & 38.18 & 35.65 & 50.26 \\
RocketKV & 0.4 & 57.77 & 39.39 & 33.91 & 50.09 \\
CriticalKV & 0.4 & 50.50&30.30&21.74&41.80 \\
LazyEviction & 0.4 & 59.77 & 39.70 & 29.57 & 50.96\\
TransAttn & 0.4 & 58.63& 43.94 & 33.04 & 51.83 \\
\cellcolor{oursrow}\textbf{PASK (Ours)} & \cellcolor{oursrow}\textbf{0.4} & \cellcolor{oursrow}\textbf{61.34} & \cellcolor{oursrow}\textbf{42.42} & \cellcolor{oursrow}\textbf{40.00} & \cellcolor{oursrow}\textbf{53.75} \\
\midrule
H2O & 0.5 & 44.37 & 21.21 & 22.61 & 35.51 \\
SnapKV & 0.5 & 55.78 & 37.27 & 19.13 & 46.77 \\
MiKV & 0.5 & 60.20&40.30&38.26&52.27 \\
RocketKV & 0.5 & 59.63&46.36&38.26&53.66\\
CriticalKV & 0.5 & 52.35&35.15&25.22&44.68 \\
LazyEviction & 0.5 & 60.49 & 43.03 & 34.78 & 52.88 \\
TransAttn & 0.5 & 60.06 & 43.03 & 39.13 & 53.05 \\
\cellcolor{oursrow}\textbf{PASK (Ours)} & \cellcolor{oursrow}\textbf{0.5} & \cellcolor{oursrow}\textbf{61.20} & \cellcolor{oursrow}\textbf{44.24} & \cellcolor{oursrow}\textbf{36.52} & \cellcolor{oursrow}\textbf{53.84} \\
\midrule
H2O & 0.6 & 50.50 & 27.27 & 23.48 & 41.10 \\
SnapKV & 0.6 & 57.06 & 40.91 & 31.30 & 49.83 \\
MiKV & 0.6 & 61.77 & 41.52 & 42.61 & 54.01 \\
RocketKV & 0.6 & 60.77 & 46.06 & 39.13 & 54.36 \\
CriticalKV & 0.6 & 58.92 & 41.52 & 29.57 & 50.96 \\
LazyEviction & 0.6 & 60.91 & 49.09 & 42.61 & 55.67\\
TransAttn & 0.6 & 61.63 & 44.55 & 42.61 & 54.80 \\
\cellcolor{oursrow}\textbf{PASK (Ours)} & \cellcolor{oursrow}\textbf{0.6} & \cellcolor{oursrow}\textbf{62.48} & \cellcolor{oursrow}\textbf{46.06} & \cellcolor{oursrow}\textbf{44.35} & \cellcolor{oursrow}\textbf{55.93} \\
\bottomrule
\end{tabular}
\caption{Detailed BIRD database-disjoint held-out execution accuracy (EX, \%) for Qwen3-4B.}
\vspace{-0.1in}
\label{tab:app-bird-difficulty}
\end{table*}

\subsection{Detailed efficiency results}

Table~\ref{tab:system-efficiency} reports TPOT and throughput across input/output lengths on A100 and L40S.

\begin{table*}[t]
\centering
\footnotesize
\setlength{\tabcolsep}{6pt}
\renewcommand{\arraystretch}{1}
\begin{tabular}{@{}c|c|cc|cc@{}}
\toprule
\multirow[c]{2}{*}[-0.5ex]{\textbf{Input / Output}}
& \multirow[c]{2}{*}[-0.5ex]{\textbf{Method}}
& \multicolumn{2}{c|}{\textbf{NVIDIA A100}}
& \multicolumn{2}{c}{\textbf{NVIDIA L40S}} \\
\cmidrule(lr){3-4}\cmidrule(lr){5-6}
& & \textbf{TPOT (ms) $\downarrow$} & \textbf{Throughput (T/S) $\uparrow$}
& \textbf{TPOT (ms) $\downarrow$} & \textbf{Throughput (T/S) $\uparrow$} \\
\midrule
\multirow[c]{4}{*}{8k / 256} & Full KV & $\mathrm{92.73}_{\mathrm{(0.0\%)}}$ & $\mathrm{79.14}_{\mathrm{(0.0\%)}}$ & $\mathrm{65.09}_{\mathrm{(0.0\%)}}$ & $\mathrm{107.26}_{\mathrm{(0.0\%)}}$ \\
& SnapKV & $\mathrm{54.66}_{\mathrm{(41.1\%)}}$ & $\mathrm{120.00}_{\mathrm{(51.6\%)}}$ & $\mathrm{56.80}_{\mathrm{(12.7\%)}}$ & $\mathrm{116.04}_{\mathrm{(8.2\%)}}$ \\
& RocketKV & $\mathrm{81.72}_{\mathrm{(11.9\%)}}$ & $\mathrm{93.04}_{\mathrm{(17.6\%)}}$ & $\mathrm{61.05}_{\mathrm{(6.2\%)}}$ & $\mathrm{103.41}_{\mathrm{(-3.6\%)}}$ \\
& CriticalKV & $\mathrm{54.34}_{\mathrm{(41.4\%)}}$ & $\mathrm{99.68}_{\mathrm{(26.0\%)}}$ & $\mathrm{49.09}_{\mathrm{(24.6\%)}}$ & $\mathrm{115.07}_{\mathrm{(7.3\%)}}$ \\
& \cellcolor{oursrow}\textbf{PASK (Ours)} & \cellcolor{oursrow}$\mathbf{45.89}_{\mathrm{(50.5\%)}}$ & \cellcolor{oursrow}$\mathbf{140.16}_{\mathrm{(77.1\%)}}$ & \cellcolor{oursrow}$\mathbf{43.62}_{\mathrm{(33.0\%)}}$ & \cellcolor{oursrow}$\mathbf{146.15}_{\mathrm{(36.3\%)}}$ \\
\midrule
\multirow[c]{4}{*}{8k / 1024} & Full KV & $\mathrm{89.49}_{\mathrm{(0.0\%)}}$ & $\mathrm{87.26}_{\mathrm{(0.0\%)}}$ & $\mathrm{63.45}_{\mathrm{(0.0\%)}}$ & $\mathrm{121.67}_{\mathrm{(0.0\%)}}$ \\
& SnapKV & $\mathrm{51.29}_{\mathrm{(42.7\%)}}$ & $\mathrm{150.04}_{\mathrm{(71.9\%)}}$ & $\mathrm{52.12}_{\mathrm{(17.9\%)}}$ & $\mathrm{145.01}_{\mathrm{(19.2\%)}}$ \\
& RocketKV & $\mathrm{116.15}_{\mathrm{(-29.8\%)}}$ & $\mathrm{68.21}_{\mathrm{(-21.8\%)}}$ & $\mathrm{66.75}_{\mathrm{(-5.2\%)}}$ & $\mathrm{95.77}_{\mathrm{(-21.3\%)}}$ \\
& CriticalKV & $\mathrm{45.03}_{\mathrm{(49.7\%)}}$ & $\mathrm{155.22}_{\mathrm{(77.9\%)}}$ & $\mathrm{41.35}_{\mathrm{(34.8\%)}}$ & $\mathrm{172.30}_{\mathrm{(41.6\%)}}$ \\
& \cellcolor{oursrow}\textbf{PASK (Ours)} & \cellcolor{oursrow}$\mathbf{43.28}_{\mathrm{(51.6\%)}}$ & \cellcolor{oursrow}$\mathbf{173.58}_{\mathrm{(98.9\%)}}$ & \cellcolor{oursrow}$\mathbf{40.15}_{\mathrm{(36.7\%)}}$ & \cellcolor{oursrow}$\mathbf{186.45}_{\mathrm{(53.2\%)}}$ \\
\midrule
\multirow[c]{4}{*}{32k / 256} & Full KV & $\mathrm{256.04}_{\mathrm{(0.0\%)}}$ & $\mathbf{25.77}_{\mathrm{(0.0\%)}}$ & $\mathrm{87.85}_{\mathrm{(0.0\%)}}$ & $\mathrm{9.72}_{\mathrm{(0.0\%)}}$ \\
& SnapKV & $\mathrm{103.72}_{\mathrm{(59.5\%)}}$ & $\mathrm{15.32}_{\mathrm{(-40.6\%)}}$ & $\mathrm{89.28}_{\mathrm{(-1.6\%)}}$ & $\mathrm{17.49}_{\mathrm{(79.9\%)}}$ \\
& RocketKV & $\mathrm{201.13}_{\mathrm{(21.4\%)}}$ & $\mathrm{13.52}_{\mathrm{(-47.5\%)}}$ & $\mathrm{85.42}_{\mathrm{(2.8\%)}}$ & $\mathrm{15.81}_{\mathrm{(62.7\%)}}$ \\
& CriticalKV & $\mathrm{117.36}_{\mathrm{(54.2\%)}}$ & $\mathrm{13.34}_{\mathrm{(-48.2\%)}}$ & $\mathrm{98.83}_{\mathrm{(-12.5\%)}}$ & $\mathrm{15.68}_{\mathrm{(61.3\%)}}$ \\
& \cellcolor{oursrow}\textbf{PASK (Ours)} & \cellcolor{oursrow}$\mathbf{84.17}_{\mathrm{(67.1\%)}}$ & \cellcolor{oursrow}$\mathrm{18.57}_{\mathrm{(-27.9\%)}}$ & \cellcolor{oursrow}$\mathbf{74.52}_{\mathrm{(15.2\%)}}$ & \cellcolor{oursrow}$\mathbf{21.26}_{\mathrm{(118.7\%)}}$ \\
\midrule
\multirow[c]{4}{*}{32k / 1024} & Full KV & $\mathrm{227.52}_{\mathrm{(0.0\%)}}$ & $\mathrm{33.22}_{\mathrm{(0.0\%)}}$ & $\mathrm{83.94}_{\mathrm{(0.0\%)}}$ & $\mathrm{31.92}_{\mathrm{(0.0\%)}}$ \\
& SnapKV & $\mathrm{89.09}_{\mathrm{(60.8\%)}}$ & $\mathrm{42.44}_{\mathrm{(27.8\%)}}$ & $\mathrm{74.27}_{\mathrm{(11.5\%)}}$ & $\mathrm{48.15}_{\mathrm{(50.8\%)}}$ \\
& RocketKV & $\mathrm{195.98}_{\mathrm{(13.9\%)}}$ & $\mathrm{35.49}_{\mathrm{(6.8\%)}}$ & $\mathrm{100.90}_{\mathrm{(-20.2\%)}}$ & $\mathrm{28.62}_{\mathrm{(-10.3\%)}}$ \\
& CriticalKV & $\mathrm{89.80}_{\mathrm{(60.5\%)}}$ & $\mathrm{38.74}_{\mathrm{(16.6\%)}}$ & $\mathrm{76.53}_{\mathrm{(8.8\%)}}$ & $\mathrm{44.60}_{\mathrm{(39.7\%)}}$ \\
& \cellcolor{oursrow}\textbf{PASK (Ours)} & \cellcolor{oursrow}$\mathbf{68.69}_{\mathrm{(69.8\%)}}$ & \cellcolor{oursrow}$\mathbf{51.02}_{\mathrm{(53.6\%)}}$ & \cellcolor{oursrow}$\mathbf{60.58}_{\mathrm{(27.8\%)}}$ & \cellcolor{oursrow}$\mathbf{50.84}_{\mathrm{(59.3\%)}}$ \\
\midrule
\multirow[c]{4}{*}{32k / 2048} & Full KV & $\mathrm{224.95}_{\mathrm{(0.0\%)}}$ & $\mathrm{34.53}_{\mathrm{(0.0\%)}}$ & $\mathrm{61.09}_{\mathrm{(0.0\%)}}$ & $\mathrm{31.49}_{\mathrm{(0.0\%)}}$ \\
& SnapKV & $\mathrm{97.58}_{\mathrm{(56.6\%)}}$ & $\mathrm{43.37}_{\mathrm{(25.6\%)}}$ & $\mathrm{73.59}_{\mathrm{(-20.5\%)}}$ & $\mathrm{51.29}_{\mathrm{(62.9\%)}}$ \\
& RocketKV & $\mathrm{150.87}_{\mathrm{(32.9\%)}}$ & $\mathrm{38.13}_{\mathrm{(10.4\%)}}$ & $\mathrm{91.90}_{\mathrm{(-50.4\%)}}$ & $\mathrm{33.17}_{\mathrm{(5.3\%)}}$ \\
& CriticalKV & $\mathrm{97.94}_{\mathrm{(56.5\%)}}$ & $\mathrm{41.64}_{\mathrm{(20.6\%)}}$ & $\mathrm{74.84}_{\mathrm{(-22.5\%)}}$ & $\mathrm{48.95}_{\mathrm{(55.4\%)}}$ \\
& \cellcolor{oursrow}\textbf{PASK (Ours)} & \cellcolor{oursrow}$\mathbf{67.28}_{\mathrm{(70.1\%)}}$ & \cellcolor{oursrow}$\mathbf{52.21}_{\mathrm{(51.2\%)}}$ & \cellcolor{oursrow}$\mathbf{54.87}_{\mathrm{(10.2\%)}}$ & \cellcolor{oursrow}$\mathbf{52.98}_{\mathrm{(68.2\%)}}$ \\
\bottomrule
\end{tabular}
\vspace{-0.1in}
\caption{Serving efficiency for Qwen3-4B at concurrency $c=8$ and approximate total KV budget $B=0.33$ across input/output lengths. Parenthetical percentages are relative to Full KV on the same GPU: positive TPOT values indicate reductions, positive throughput values indicate increases, and negative values indicate regressions.}
\vspace{-0.1in}
\label{tab:system-efficiency}
\end{table*}


\subsection{High-concurrency scaling}

Table~\ref{tab:app-system-scaling} extends the main system results to higher concurrency at the 1024-token checkpoint. It focuses on 32K contexts, where PASK is designed to reduce persistent KV traffic during concurrent decode.

\begin{table*}[t]
\centering
\footnotesize
\setlength{\tabcolsep}{4pt}
\renewcommand{\arraystretch}{1.15}
\begin{tabular}{@{}llr|rrr|rrr@{}}
\toprule
\multirow[c]{2}{*}[-0.5ex]{\textbf{GPU}} & \multirow[c]{2}{*}[-0.5ex]{\textbf{Input}} & \multirow[c]{2}{*}[-0.5ex]{\textbf{\(c\)}} & \multicolumn{3}{c|}{\textbf{Throughput (tokens/s) $\uparrow$}} & \multicolumn{3}{c}{\textbf{TPOT (ms) $\downarrow$}} \\
\cmidrule(lr){4-6}\cmidrule(lr){7-9}
 & & & \textbf{Full KV} & \textbf{PASK} & \textbf{SnapKV} & \textbf{Full KV} & \textbf{PASK} & \textbf{SnapKV} \\
\midrule
A100 & 32K & 8  & $\mathrm{33.22}_{\mathrm{(0.0\%)}}$ & \cellcolor{oursrow}$\mathbf{51.02}_{\mathrm{(53.6\%)}}$ & $\mathrm{42.44}_{\mathrm{(27.8\%)}}$ & $\mathrm{227.52}_{\mathrm{(0.0\%)}}$ & \cellcolor{oursrow}$\mathbf{68.69}_{\mathrm{(69.8\%)}}$ & $\mathrm{89.09}_{\mathrm{(60.8\%)}}$ \\
A100 & 32K & 16 & $\mathrm{33.26}_{\mathrm{(0.0\%)}}$ & \cellcolor{oursrow}$\mathbf{63.23}_{\mathrm{(90.1\%)}}$ & $\mathrm{51.84}_{\mathrm{(55.9\%)}}$ & $\mathrm{331.35}_{\mathrm{(0.0\%)}}$ & \cellcolor{oursrow}$\mathbf{68.00}_{\mathrm{(79.5\%)}}$ & $\mathrm{140.50}_{\mathrm{(57.6\%)}}$ \\
A100 & 32K & 32 & $\mathrm{33.34}_{\mathrm{(0.0\%)}}$ & \cellcolor{oursrow}$\mathbf{72.23}_{\mathrm{(116.6\%)}}$ & $\mathrm{51.66}_{\mathrm{(54.9\%)}}$ & $\mathrm{333.99}_{\mathrm{(0.0\%)}}$ & \cellcolor{oursrow}$\mathbf{68.96}_{\mathrm{(79.4\%)}}$ & $\mathrm{141.81}_{\mathrm{(57.5\%)}}$ \\
L40S & 32K & 8  & $\mathrm{31.92}_{\mathrm{(0.0\%)}}$ & \cellcolor{oursrow}$\mathbf{50.84}_{\mathrm{(59.3\%)}}$ & $\mathrm{48.15}_{\mathrm{(50.8\%)}}$ & $\mathrm{83.94}_{\mathrm{(0.0\%)}}$ & \cellcolor{oursrow}$\mathbf{60.58}_{\mathrm{(27.8\%)}}$ & $\mathrm{74.27}_{\mathrm{(11.5\%)}}$ \\
L40S & 32K & 16 & $\mathrm{32.02}_{\mathrm{(0.0\%)}}$ & \cellcolor{oursrow}$\mathbf{60.38}_{\mathrm{(88.6\%)}}$ & $\mathrm{48.29}_{\mathrm{(50.8\%)}}$ & $\mathrm{85.54}_{\mathrm{(0.0\%)}}$ & \cellcolor{oursrow}$\mathbf{60.74}_{\mathrm{(29.0\%)}}$ & $\mathrm{73.96}_{\mathrm{(13.5\%)}}$ \\
L40S & 32K & 32 & $\mathrm{32.50}_{\mathrm{(0.0\%)}}$ & \cellcolor{oursrow}$\mathbf{71.37}_{\mathrm{(119.6\%)}}$ & $\mathrm{48.09}_{\mathrm{(48.0\%)}}$ & $\mathrm{85.41}_{\mathrm{(0.0\%)}}$ & \cellcolor{oursrow}$\mathbf{61.78}_{\mathrm{(27.7\%)}}$ & $\mathrm{74.32}_{\mathrm{(13.0\%)}}$ \\
\bottomrule
\end{tabular}
\vspace{-0.08in}
\caption{High-concurrency system results for 32K inputs at the 1024-token checkpoint. Each cell is the median over three successful repetitions. Parenthetical percentages are relative to Full KV on the same GPU: positive throughput values indicate increases and positive TPOT values indicate reductions.}
\label{tab:app-system-scaling}
\vspace{-0.08in}
\end{table*}

Across \(c\in\{8,16,32\}\), PASK increases throughput and reduces TPOT relative to Full KV on both GPUs.

\subsection{Reliability-tolerance sensitivity}
\label{app:alpha-selection}

\begin{figure}[t]
\centering
\includegraphics[width=1.0\linewidth, trim=0mm 0mm 0mm 2mm, clip]{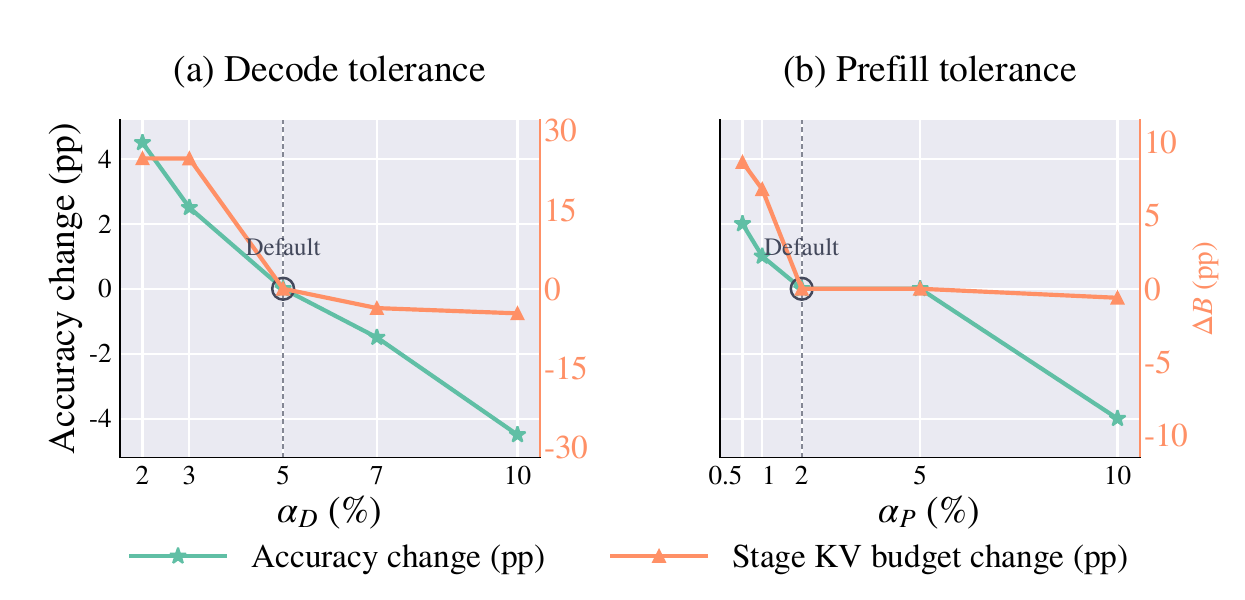}
\vspace{-0.4in}
\caption{Development-set sensitivity to (a) Decode tolerance $\alpha_{\mathrm D}$ and (b) Prefill tolerance $\alpha_{\mathrm P}$. Accuracy and stage-budget changes are measured in percentage points relative to the respective defaults $5\%$ and $2\%$.}
\label{fig:stagewise-reliability-cost}
\vspace{-0.1in}
\end{figure}

For each stage, we select the lowest-budget candidate satisfying the development reliability gate. Fig~\ref{fig:stagewise-reliability-cost} shows the resulting accuracy--budget trade-off around the defaults $\alpha_{\mathrm D}=5\%$ and $\alpha_{\mathrm P}=2\%$.

\subsection{Calibration Cost}
\label{app:calibration-cost}

Table~\ref{tab:calibration-cost} reports the measured offline calibration cost on a single NVIDIA A100-80GB GPU (Qwen3-4B, the BFCL non-Live calibration split of 200 cases per candidate, the same side table and runtime contract that produced the deployed policy). Runtime-V perturbs one structure--layer bucket at a time and Runtime-D measures the fixed-trajectory distortion of each candidate; median candidate runtimes are 218.8 s for Runtime-V and 222.1 s for Runtime-D. With two concurrent workers sharing the same GPU, the full 54-candidate scans take 1.64 and 1.67 GPU-hours, respectively. The combined Runtime-V and Runtime-D calibration takes approximately 3.3 GPU-hours on one A100. The calibration is therefore a one-time cost of a few GPU-hours, amortized over all downstream use of the frozen lookup tables.

\begin{table}[t]
\centering
\footnotesize
\setlength{\tabcolsep}{4pt}
\renewcommand{\arraystretch}{1.03}
\begin{tabular}{@{}lcc@{}}
\toprule
& \textbf{Runtime-V} & \textbf{Runtime-D} \\
\midrule
All-high control & 220.4\,s & 260.5\,s \\
Candidate time, median & 218.8\,s & 222.1\,s \\
Candidate time, p90 & 221.1\,s & 226.5\,s \\
GPU util, mean / peak & \multicolumn{2}{c}{62\% / 100\%} \\
Peak memory & \multicolumn{2}{c}{75.8 GiB} \\
Full 54-candidate scan & 1.64 h & 1.67 h \\
\bottomrule
\end{tabular}
\captionsetup{skip=2pt}
\caption{Measured calibration cost on one NVIDIA A100-80GB (Qwen3-4B, BFCL non-Live calibration split, 200 cases per candidate).}
\label{tab:calibration-cost}
\end{table}

\subsection{Direct Evidence of Responsibility--Distortion Mismatch}
\label{app:mismatch-evidence}

We compare task responsibility and model-side numerical sensitivity on the calibration buckets used to construct the frozen policy. For each action $a\in\{R,L\}$, task responsibility is measured by the effective sensitivity $\bar V(q,a)$ from Eq.~\ref{eq:task-sensitivity}, while model-side sensitivity is measured by the corresponding distortion $D(q,a)$ from Eq.~\ref{eq:distortion}.
For compactness, let
\[
\mathcal{T}^{(a)}
=
\operatorname{Top}_{20\%}\!\left(\bar V(\cdot,a)\right).
\]

\begin{equation}
\begin{aligned}
\mathrm{Recall}^{(a)}_{20\%}(S)
&=
\frac{
\left|
\mathcal{T}^{(a)}
\cap
\operatorname{Top}_{20\%}\!\left(S(\cdot,a)\right)
\right|
}{
\left|\mathcal{T}^{(a)}\right|
}.
\end{aligned}
\end{equation}

\begin{equation}
\begin{aligned}
\mathrm{Mismatch}^{(a)}_{20,50}(S)
&=
\frac{
\left|
\mathcal{T}^{(a)}
\cap
\operatorname{Bottom}_{50\%}\!\left(S(\cdot,a)\right)
\right|
}{
\left|\mathcal{T}^{(a)}\right|
}.
\label{eq:mismatch-rate}
\end{aligned}
\end{equation}
A high mismatch value means that the model-side ranking places many task-critical buckets in its lower half. The attention-saliency quantity $A(q)$ requires a separate reference-attention pass and is not inferred from $D(q,a)$.

\begin{table}[t]
\centering
\footnotesize
\setlength{\tabcolsep}{2.5pt}
\renewcommand{\arraystretch}{1.08}
\begin{tabular}{@{}lccc@{}}
\toprule
\textbf{Action} & \textbf{\shortstack{Model-side\\score}} & \textbf{\shortstack{Recall\\@20\%}} & \textbf{\shortstack{Mismatch\\$_{20,50}$}} \\
\midrule
Release $R$ & $D(q,R)$ & $0.0\%$ & $100.0\%$ \\
Retain-low $L$ & $D(q,L)$ & $0.0\%$ & $83.3\%$ \\
\bottomrule
\end{tabular}
\caption{Direct comparison between effective task responsibility and Runtime-D distortion on 27 Qwen3-4B BFCL non-live calibration buckets. The effective responsibility values use the monotone envelope in Eq.~\ref{eq:task-sensitivity}.}
\label{tab:mismatch-correlation}
\end{table}

\begin{figure}[t]
\centering
\includegraphics[width=0.5\textwidth]{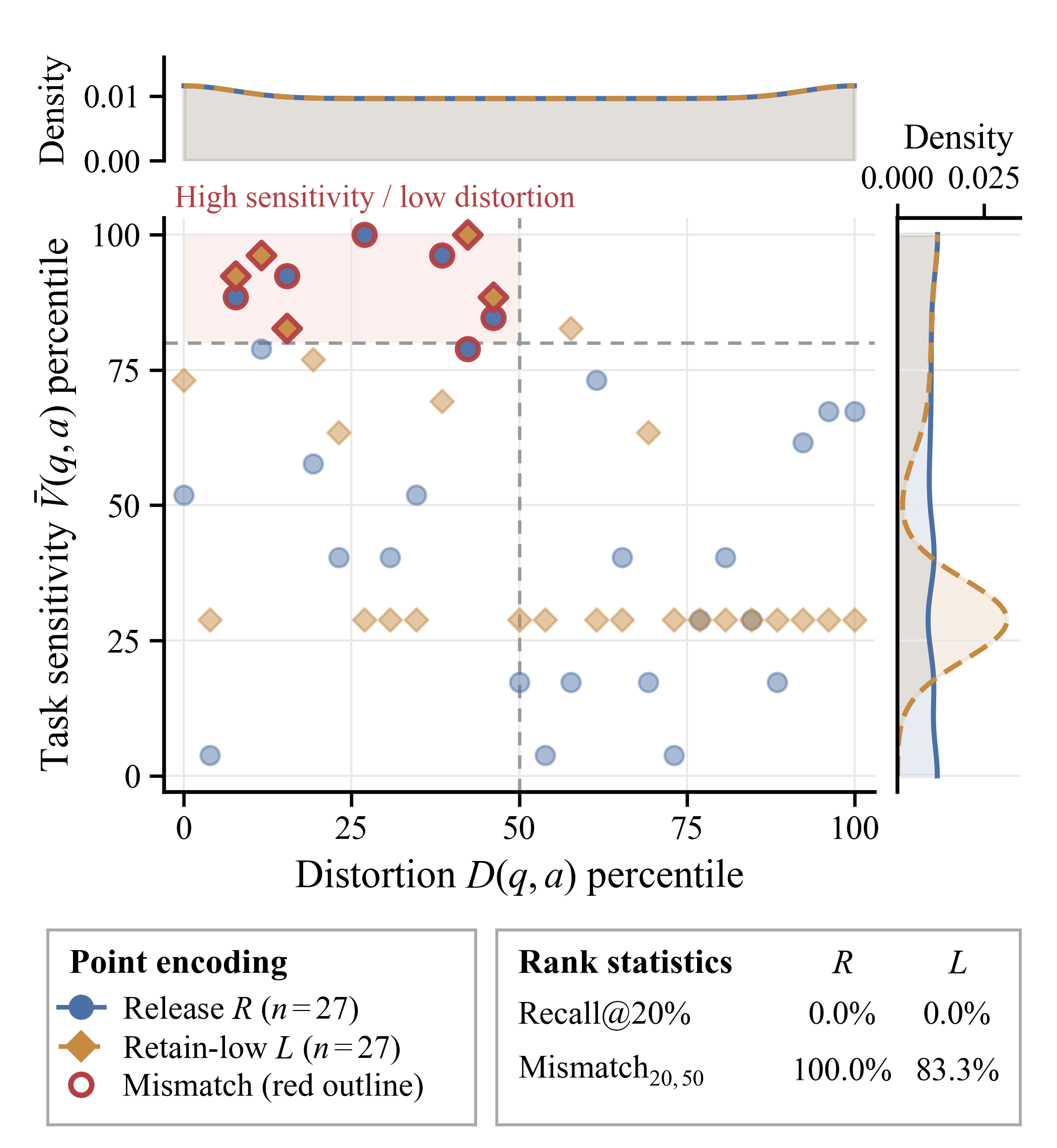}
\caption{Responsibility--distortion mismatch across 27 calibrated structure--layer buckets. Blue circles represent release ($R$), and orange diamonds represent retain-low ($L$). Axes show percentile ranks of distortion $D(q,a)$ and task sensitivity $\bar V(q,a)$. Red outlines highlight top-20\% sensitivity buckets ranked in the bottom 50\% by distortion. Top and right curves show marginal kernel density estimates.}
\label{fig:mismatch-scatter}
\end{figure}
Across the 27 calibration buckets, the top-20\% sets ranked by task-error sensitivity $\bar V(q,a)$ and distortion $D(q,a)$ do not overlap for either action (Table~\ref{tab:mismatch-correlation}). 

\subsection{Prefill--Decode Decomposition}
\label{app:prefill-decode-decomposition}

We decompose the policy into four prefill--decode combinations under a single strict JSON-schema BFCL non-live test protocol. The prefill policy was selected on development data using the $\alpha_{\mathrm{P}}=2\%$ gate with decode fixed to all-high; the decode policy was frozen to the default $\tau=\{0.5\%,1\%\}$. This stage decomposition measures each policy's contribution and their composition.

\begin{table}[t]
\centering
\scriptsize
\setlength{\tabcolsep}{2pt}
\renewcommand{\arraystretch}{1.06}
\begin{tabular}{@{}lcc|ccc|c@{}}
\toprule
\textbf{Variant} & \textbf{\shortstack{Prefill\\policy}} & \textbf{\shortstack{Decode\\policy}} & \textbf{\(B_{\mathrm P}^{\mathrm{meas}}$ $\downarrow$} & \textbf{\(B_{\mathrm D}^{\mathrm{meas}}$ $\downarrow$} & \textbf{\shortstack{Total\\\(B\) $\downarrow$}} & \textbf{\shortstack{BFCL non-live\\Acc. $\uparrow$}} \\
\midrule
Full KV & Full & Full & 1.000 & 1.000 & 1.000 & 93.00 \\
Prefill-only & PASK  & Full & 0.250 & 1.000 & 0.489 & 91.17 \\
Decode-only & Full & PASK  & 1.000 & 0.501 & 0.841 & 90.67 \\
Combined PASK & PASK  & PASK  & 0.250 & 0.501 & 0.330 & 88.83 \\
\bottomrule
\end{tabular}
\caption{Prefill--decode stage decomposition for Qwen3-14B on BFCL non-live under the strict JSON-schema protocol.}
\label{tab:prefill-decode-decomposition}
\end{table}

\textbf{Protocol and budget accounting.} All four rows use BFCL non-live test cases at concurrency 1 with \texttt{max\_new\_tokens}=192. The Full KV and Decode-only rows reuse the matching strict-protocol controls; Prefill-only and Combined use the development-frozen minimum-protection prefill policy. Following Section~\ref{sec:exp-setup}, we report measured KV costs normalized by the corresponding Full-KV reference. $B_{\mathrm P}^{\mathrm{meas}}$ and $B_{\mathrm D}^{\mathrm{meas}}$ denote the prefill and decode cost ratios, respectively. Total $B$ is computed as the sum of measured prefill and decode KV payloads divided by the sum of their corresponding Full-KV reference payloads.

\textbf{Result.} Prefill-only retained 91.17\% accuracy at total $B=0.489$, whereas decode-only retained 90.67\% at total $B=0.841$. Activating both stages yielded 88.83\% accuracy at total $B=0.330$, a 4.17-percentage-point difference from Full KV. These operating points are not matched in total budget, so the table reports stage contributions under the frozen policies.
\section{Theoretical Properties}
\label{app:theoretical-properties}

We formalize properties of the floor-constrained controller, greedy calibration path, online complexity, and empirical reserve construction.

\subsection{Protection-floor preservation}

\begin{proposition}[Protection floors are invariant]
\label{prop:floor-preservation}
Assume \textup{\textsc{Demote}} in Algorithm~\ref{alg:pask-online} only applies an adjacent demotion when the resulting action remains no lower than the stored floor $a^-_{t,c}$, as specified in Section~\ref{sec:online-lookup}. Then for every accepted token $t$ and layer group $c$, the persisted action satisfies
\begin{equation}
a_{t,c}\succeq a^-_{t,c}.
\end{equation}
\end{proposition}

\begin{proof}
Algorithm~\ref{alg:pask-online} initializes
\[
a_{t,c}
=\max_{\preceq}(a^0_{t,c},a^-_{t,c}),
\]
which satisfies the invariant. By assumption, \textsc{Demote} cannot cross the floor, while \textsc{Promote} only moves upward in the order $R\prec L\prec H$. Both operations therefore preserve the invariant, which holds for the final persisted action.
\end{proof}
This invariant guarantees floor preservation, not task correctness; the latter depends on the empirical calibration in Eq.~\ref{eq:task-sensitivity}.
\subsection{Nested residual-allocation frontier}
\begin{proposition}[Nested greedy path]

Fix a threshold pair $\tau$ and its floor policy $a_{\min}^{\tau}$. Let $\Theta_0,\Theta_1,\ldots,\Theta_J$ denote the bucket-action maps produced by \textsc{GreedyPath}, where each step applies one eligible adjacent upgrade $R\to L$ or $L\to H$ without demotion.

For fixed reference trajectories $\mathcal{D}_{\mathrm{ref}}$, define the nominal reference-trajectory KV cost as
\[
\widetilde{C}_{\mathrm{KV}}^{\mathrm{ref}}(\Theta)
=
\frac{
\sum_{r\in\mathcal{D}_{\mathrm{ref}}}
\sum_{t=1}^{T_r^{\mathrm{ref}}}
\sum_{c\in\mathcal{C}}
m\!\left(\Theta(q_{r,t,c}^{\mathrm{ref}}),c\right)
}{
C_{\mathrm{full}}
\sum_{r\in\mathcal{D}_{\mathrm{ref}}}
T_r^{\mathrm{ref}}
}.
\]
Here, $q_{r,t,c}^{\mathrm{ref}}$ denotes the structure--layer bucket of a token on the fixed reference trajectory.

Then the action maps are componentwise ordered:
\begin{equation}
\Theta_0 \preceq \Theta_1
\preceq \cdots \preceq \Theta_J.
\end{equation}
Their nominal reference-trajectory costs are nondecreasing:
\begin{equation}
\widetilde{C}_{\mathrm{KV}}^{\mathrm{ref}}(\Theta_0)
\leq
\widetilde{C}_{\mathrm{KV}}^{\mathrm{ref}}(\Theta_1)
\leq \cdots \leq
\widetilde{C}_{\mathrm{KV}}^{\mathrm{ref}}(\Theta_J).
\end{equation}
\end{proposition}

\begin{proof}
By construction, each greedy step upgrades one bucket action without demoting any other action. Thus, $\Theta_i\preceq\Theta_{i+1}$ componentwise. Since the reference trajectories and their bucket assignments are fixed, and $m(a',c)>m(a,c)$ for each adjacent upgrade $a\prec a'$, the nominal payload numerator cannot decrease. The normalization is also fixed, establishing the cost inequality.
\end{proof}
The proposition applies to each fixed-$\tau$ greedy path. Candidate paths generated from different floor thresholds in Algorithm 1 need not be totally ordered, and development selection evaluates them explicitly using Eq. 9. The monotonicity result concerns nominal costs evaluated on fixed reference trajectories, rather than the realized autoregressive cost defined in Eq. 2. During deployment, policy-dependent generation trajectories and budget-controller adjustments may change realized costs, for which monotonicity is not guaranteed.

\subsection{Online computational complexity}

\begin{proposition}[History-length-independent online control]
\label{prop:online-complexity}
Let $|\mathcal C|$ be the number of layer groups and assume constant-time lookup of one frozen table entry. For each accepted token, Algorithm~\ref{alg:pask-online} requires $O(|\mathcal C|)$ lookup/storage work and at most $O(|\mathcal C|\log|\mathcal C|)$ work to rank the $O(|\mathcal C|)$ eligible promotions or demotions. Its request-local controller state is $O(|\mathcal C|)$. Consequently, for the fixed three-group implementation used in this work, the online control overhead is $O(1)$ with respect to sequence length and retained-history length.
\end{proposition}

\begin{proof}
The lookup key is formed once per layer group, giving $|\mathcal C|$ table accesses and an action vector of length $|\mathcal C|$. Each group has only the three ordered actions $R\prec L\prec H$, so at a token there are only $O(|\mathcal C|)$ adjacent promotion/demotion candidates. Ranking these candidates by their stored $\rho_D$ scores is bounded by $O(|\mathcal C|\log|\mathcal C|)$ using a comparison sort, and applying the selected adjustments is linear in $|\mathcal C|$. The controller stores only the action/floor/score vectors and scalar budget state. None of these quantities grows with the number of previously generated tokens.
\end{proof}

This proposition concerns the policy-controller work only; the Transformer attention computation still depends on the retained KV representation and the serving kernel.

\subsection{Empirical reserve coverage}

\begin{proposition}[Calibration-sample coverage of reserve deficits]
\label{prop:reserve-coverage}
For a parser state $s$ with $n_s>0$, let
\begin{equation}
\mathcal I_s=\{(r,u):s^{\mathrm{ref}}_{r,u}=s\},
\qquad n_s=|\mathcal I_s|,
\end{equation}
and define the empirical CDF of the calibration-sample deficits in Eq.~\ref{eq:reserve-prefix-deficit} as
\begin{equation}
\widehat F_{s;K,B_{\mathrm D}}(x)
=
\frac{1}{n_s}
\sum_{(r,u)\in\mathcal I_s}
\mathbb I\!\left[
S^{(K)}_{r,u}(B_{\mathrm D})\leq x
\right].
\end{equation}
If the state-conditioned reserve in Eq.~\ref{eq:reserve} uses the left empirical $\beta$-quantile,
\begin{equation}
\Gamma(s;B_{\mathrm D})
=
\inf\left\{
x:\widehat F_{s;K,B_{\mathrm D}}(x)\geq\beta
\right\},
\end{equation}
then
\begin{equation}
\widehat F_{s;K,B_{\mathrm D}}\!\left(\Gamma(s;B_{\mathrm D})\right)
\geq\beta,
\end{equation}
or equivalently, the empirical fraction of calibration deficits exceeding the reserve is at most $1-\beta$.
\end{proposition}

\begin{proof}
By definition, $\Gamma(s;B_{\mathrm D})$ is the left empirical $\beta$-quantile of the observed values $S^{(K)}_{r,u}(B_{\mathrm D})$ for state $s$. Therefore, the empirical CDF at this quantile is at least $\beta$. The empirical fraction of deficits exceeding the reserve is
$1-\widehat F_{s;K,B_{\mathrm D}}(\Gamma(s;B_{\mathrm D}))$, which is at most $1-\beta$.
\end{proof}

This result describes empirical calibration-sample coverage of the deficit statistic used to construct the reserve. It does not guarantee runtime budget feasibility or coverage on unseen or shifted trajectories. For an unseen state, the controller uses the pooled calibration quantile; the same empirical statement applies to that pooled calibration sample.

The default policies use $\beta=0.8$ with future horizon $K=8$. 

\section{Sample-level Compression Case Studies}
\label{app:sample-compression-cases}
We present one BIRD SQL request and one BFCL parallel-multiple request to illustrate how PASK assigns persistence by token role and layer group.

\begin{tcolorbox}[breakable,width=\linewidth,title={BIRD: percentage of carcinogenic molecules},colback=gray!3,colframe=gray!60,boxrule=0.4pt,fonttitle=\normalsize\bfseries,left=3pt,right=3pt,top=3pt,bottom=3pt]
\normalsize\sloppy
\textbf{Question (ID 227).} What is the percentage of molecules that are carcinogenic? Please provide your answer as a percentage with three decimal places.

\textbf{PASK output.}
{\ttfamily\raggedright
\detokenize{SELECT CAST(COUNT(CASE WHEN `label` = '+' THEN 1 END) AS FLOAT) * 100.0 / COUNT(*) AS percent FROM `molecule`;}
\par}

\textbf{Execution.} On the BIRD database, this query returns 44.3149\%; the reference query rounds the same result to 44.315\%.

\textbf{Compression trace.} The policy mapping assigns 320 \textsc{retain-low} and 84 \textsc{release} actions to the 404 prompt tokens in each layer group, with no \textsc{retain-high} actions. For the 37 non-stop SQL tokens, the reported decode trace gives the following split:
\begin{center}
\setlength{\tabcolsep}{3pt}
\begin{tabular}{@{}llrrr@{}}
\toprule
\textbf{Stage} & \textbf{Layer} & \textbf{H} & \textbf{L} & \textbf{R} \\
\midrule
Prefill & Lower & 0 & 320 & 84 \\
& Middle & 0 & 320 & 84 \\
& Upper & 0 & 320 & 84 \\
Decode & Lower & 37 & 0 & 0 \\
& Middle & 24 & 6 & 7 \\
& Upper & 3 & 34 & 0 \\
\bottomrule
\end{tabular}
\end{center}
The seven middle-group releases are pieces of the \texttt{'+'} predicate literal, the \texttt{1} case result and the \texttt{100.0} multiplier. Their lower-layer KV remains BF16 and upper-layer KV remains INT4, so each expression is retained in other groups.
\end{tcolorbox}

\begin{tcolorbox}[breakable,width=\linewidth,title={BFCL: parallel-multiple tool calls (parallel\_multiple\_183)},colback=gray!3,colframe=gray!60,boxrule=0.4pt,fonttitle=\normalsize\bfseries,left=3pt,right=3pt,top=3pt,bottom=3pt]
\normalsize\sloppy
\textbf{Input.} Could you first calculate the probability of drawing a heart from a deck of 52 cards where there are 13 hearts, and then calculate the probability of drawing a queen from the same deck where there are 4 queens? After that, could you retrieve the most recent artwork by the artist named `Pablo Picasso' with a detailed description? Finally, could you locate the most popular sculpture exhibitions in New York, NY that are happening in the month of December and have high user ratings?

\textbf{PASK output.} Four calls match the reference functions and arguments:
\begin{enumerate}[leftmargin=*]
\item \texttt{card\_game\_probability.calculate}: total\_cards=52, desired\_cards=13, cards\_drawn=1.
\item \texttt{card\_game\_probability.calculate}: total\_cards=52, desired\_cards=4, cards\_drawn=1.
\item \texttt{get\_sculpture\_info}: \texttt{artist\_name="Pablo Picasso", detail=true}.
\item \texttt{find\_exhibition}: \texttt{location="New York, NY", art\_form=sculpture, month="December", user\_ratings="high"}.
\end{enumerate}

\textbf{Prefill trace.} The matcher labels the 595 prompt tokens as protected content (181), schema scaffold (130), function names (19), descriptions (167), types (36), parameter keys (53) and enums (9). The layer-group actions are:
\begin{center}
\setlength{\tabcolsep}{3pt}
\begin{tabular}{@{}llrrr@{}}
\toprule
\textbf{Stage} & \textbf{Layer} & \textbf{H} & \textbf{L} & \textbf{R} \\
\midrule
Prefill & Lower & 81 & 181 & 333 \\
& Middle & 81 & 348 & 166 \\
& Upper & 81 & 348 & 166 \\
Decode & Lower & 143 & 0 & 0 \\
& Middle & 0 & 84 & 59 \\
& Upper & 30 & 112 & 1 \\
\bottomrule
\end{tabular}
\end{center}
The 59 middle-group decode releases comprise 25 function-name pieces, 15 optional-field-key pieces, 14 structural separators and 5 optional-value pieces. The function names remain BF16 in the lower and upper groups; released keys and separators remain in the lower group and mostly at INT4 in the upper group. The four-call output exactly matches the reference on this one case.
\end{tcolorbox}

\paragraph{Mixed-Mode Generation}
The cases above illustrate PASK within a single constrained response. Extending this control to agents that alternate between free-form language and structured calls could treat mode boundaries as structural events. The serving stack exposes when a grammar is active, allowing dispatch tokens after free-form text to receive lookahead-critical protection and free-form spans to use a dedicated mode tag with calibrated floors and upgrade priorities. The request-local budget ledger could continue across mode boundaries; in later turns, earlier free-form content becomes prompt text handled by the prefill companion (Section~\ref{sec:prefill-companion}). Cross-mode calibration would measure how interventions on free-form history affect subsequent structured calls, using jointly intervened rollouts or paired trajectories. Evaluation covering both modes would establish the accuracy--cost trade-off of this extension.





\end{document}